\documentclass[preprints,article,accept,moreauthors,dvi2pdf]{Definitions/mdpi}

\usepackage{lineno,hyperref}

\usepackage{cleveref} % clever reference package https://ctan.org/pkg/cleveref
\usepackage{colortbl} %in process table https://ctan.org/pkg/colortbl
\usepackage{booktabs}% http://ctan.org/pkg/booktabs
\usepackage{multirow} %https://ctan.org/pkg/multirow
\usepackage{makecell} %https://www.ctan.org/pkg/makecell
\usepackage{soul} 
\usepackage{tikz}
\usepackage[T1]{fontenc}
\usepackage{lmodern}

\definecolor{tableLineOne}{rgb}{0.961, 0.961, 0.961}
\definecolor{tableLineTwo}{rgb}{0.878, 0.878, 0.878}
\definecolor{DaimlerRed}{rgb}{0.443, 0.094, 0.047}
\definecolor{DaimlerPetrol}{rgb}{0, 0.404, 0.498}

\newcolumntype{M}[1]{>{\centering\arraybackslash}m{#1}}
\newcolumntype{N}{@{}m{0pt}@{}}
\newcommand\VRule[1][\arrayrulewidth]{\vrule width #1}

\firstpage{1} 
\pubvolume{1}
\issuenum{1}
\articlenumber{0}
\pubyear{2021}
\copyrightyear{2020}
\datereceived{1 March 2021} 
\dateaccepted{} 
\datepublished{} 
\hreflink{https://doi.org/} % If needed use \linebreak
\Title{Towards CRISP-ML(Q): A Machine Learning Process Model with Quality Assurance Methodology}

\TitleCitation{Towards CRISP-ML(Q): A Machine Learning Process Model with Quality Assurance Methodology}

\Author{Stefan Studer $^{1,*}$\orcidA{}, Thanh~Binh~Bui $^{2,*}$, Christian Drescher$^{1}$, Alexander Hanuschkin$^{3}$, Ludwig Winkler $^{2}$, Steven~Peters  $^{1}$\orcidB{} and Klaus-Robert~M\"uller $^{4}$\orcidC{}}

\AuthorNames{Stefan Studer, Thanh Binh Bui, Christian Drescher, Alexander Hanuschkin, Ludwig Winkler, Steven Peters and Klaus-Robert M\"uller}

\AuthorCitation{Studer, S.; Bui, T.B.; Drescher, C.; Hanuschkin, A.; Winkler, L.; Peters, S.; M\"uller, K.-R.;}
\address{%
$^{1}$ \quad \mbox{Mercedes-Benz AG}, Group Research, \mbox{Artificial~Intelligence~Research}, 71059 Sindelfingen, Germany; stefan.studer@daimler.com (S.St.); christian.d.drescher@daimler.com (C.D.); steven.peters@daimler.com (S.P.)\\
$^{2}$ \quad Machine Learning Group, Technische Universit\"at Berlin, 10587 Berlin, Germany; bui@tu-berlin.de (T.B.B.); winkler@tu-berlin.de (L.W.)\\
$^{3}$ \quad Mercedes-Benz AG, Group Research, Artificial Intelligence Research, Sindelfingen, Germany and Esslingen University of Applied Sciences, Germany;  alexander.hanuschkin@daimler.com (A.H.)\\
$^{4}$ \quad Machine Learning Group, Technische Universit\"at Berlin, 10587 Berlin, Germany and Google Research, Brain team, Berlin, Germany and Dept. of Artificial Intelligence, Korea University, Seoul 136-713, South Korea and Max-Planck-Institut f\"ur Informatik, 66123 Saarbr\"ucken, Germany; klaus.r.mueller@googlemail.com (K.-R.M.)
}

\corres{Correspondence: stefan.studer@daimler.com (S.St.) and bui@tu-berlin.de (T.B.B.); equal contribution}

\abstract{Machine learning is an established and frequently used technique in industry and academia but a standard process model to improve success and efficiency of machine learning applications is still missing. Project organizations and machine learning practitioners have a need for guidance throughout the life cycle of a machine learning application to meet business expectations. We therefore propose a process model for the development of machine learning applications, that covers six phases from defining the scope to maintaining the deployed machine learning application. The first phase combines business and data understanding as data availability oftentimes affects the feasibility of the project. The sixth phase covers state-of-the-art approaches for monitoring and maintenance of a machine learning applications, as the risk of model degradation in a changing environment is eminent. With each task of the process, we propose quality assurance methodology that is suitable to adress challenges in machine learning development that we identify in form of risks. The methodology is drawn from practical experience and scientific literature and has proven to be general and stable. The process model expands on CRISP-DM, a data mining process model that enjoys strong industry support but lacks to address machine learning specific tasks. Our work proposes an industry and application neutral process model tailored for machine learning applications with focus on technical tasks for quality assurance.}


\begin{document}
%%%%%%%%%%%%%%%%%%%%%%%%%%%%%%%%%%%%%%%%%%
%\setcounter{section}{-1} %% Remove this when starting to work on the template.
%\section{How to Use this Template}
%
%The template details the sections that can be used in a manuscript. Note that the order and names of article sections may differ from the requirements of the journal (e.g., the positioning of the Materials and Methods section). Please check the instructions on the authors' page of the journal to verify the correct order and names. For any questions, please contact the editorial office of the journal or support@mdpi.com. For LaTeX-related questions please contact latex@mdpi.com.
%%The order of the section titles is: Introduction, Materials and Methods, Results, Discussion, Conclusions for these journals: aerospace,algorithms,antibodies,antioxidants,atmosphere,axioms,biomedicines,carbon,crystals,designs,diagnostics,environments,fermentation,fluids,forests,fractalfract,informatics,information,inventions,jfmk,jrfm,lubricants,neonatalscreening,neuroglia,particles,pharmaceutics,polymers,processes,technologies,viruses,vision

\section{Introduction}\label{sec:introduction}
Many industries, such as manufacturing~\cite{lee2015cyber, brettel2014virtualization}, personal transportation~\cite{dikmen2016autonomous} and healthcare~\cite{kourou2015cancermachinelearning, esteva2017dermatologist} are currently undergoing a process of digital transformation, challenging established processes with machine learning-driven approaches. The expanding demand is highlighted by the Gartner report~\cite{anha19}, claiming that organizations expect to double the number of machine learning (ML) projects within a year.

However, 75 to 85 percent of practical ML projects currently do not match their sponsors' expectations, according to surveys of leading technology companies~\cite{pactera19}. \citet{make3010004} name data and software quality among others as the key challenges in the machine learning life cycle. Another reason is the lack of guidance through standards and development process models specific to ML applications. Industrial organizations, in particular, rely heavily on standards to guarantee a consistent quality of their products or services. 
A Japanese industry Consortium (QA4AI) was founded to address those needs~\cite{hamada2020}.
%A selection of the Japanese industry founded the QA4AI Consortium to address those needs \cite{hamada2020}.

Due to the lack of a process model for ML applications, many project organizations rely on alternative models that are closely related to ML, such as, the Cross-Industry Standard Process model for Data Mining~(CRISP-DM)~\cite{crisp-dm,crisp-dm:hipp,crisp-dm:shearer}. It is grounded on industrial data mining experience~\cite{crisp-dm:shearer} and is considered most suitable for industrial projects amongst related process models~\cite{kumu06}. In fact, CRISP-DM has become the de-facto industry standard~\cite{Mariscal_2010} process model for data mining, with an expanding number of applications~\cite{krbokrprsczi07}, e.g., in quality diagnostics~\cite{abdilocu04}, marketing~\cite{gewiar00}, and warranty~\cite{hili99}. 

\begin{figure}
	\centering
	\includegraphics[width=0.45\textwidth]{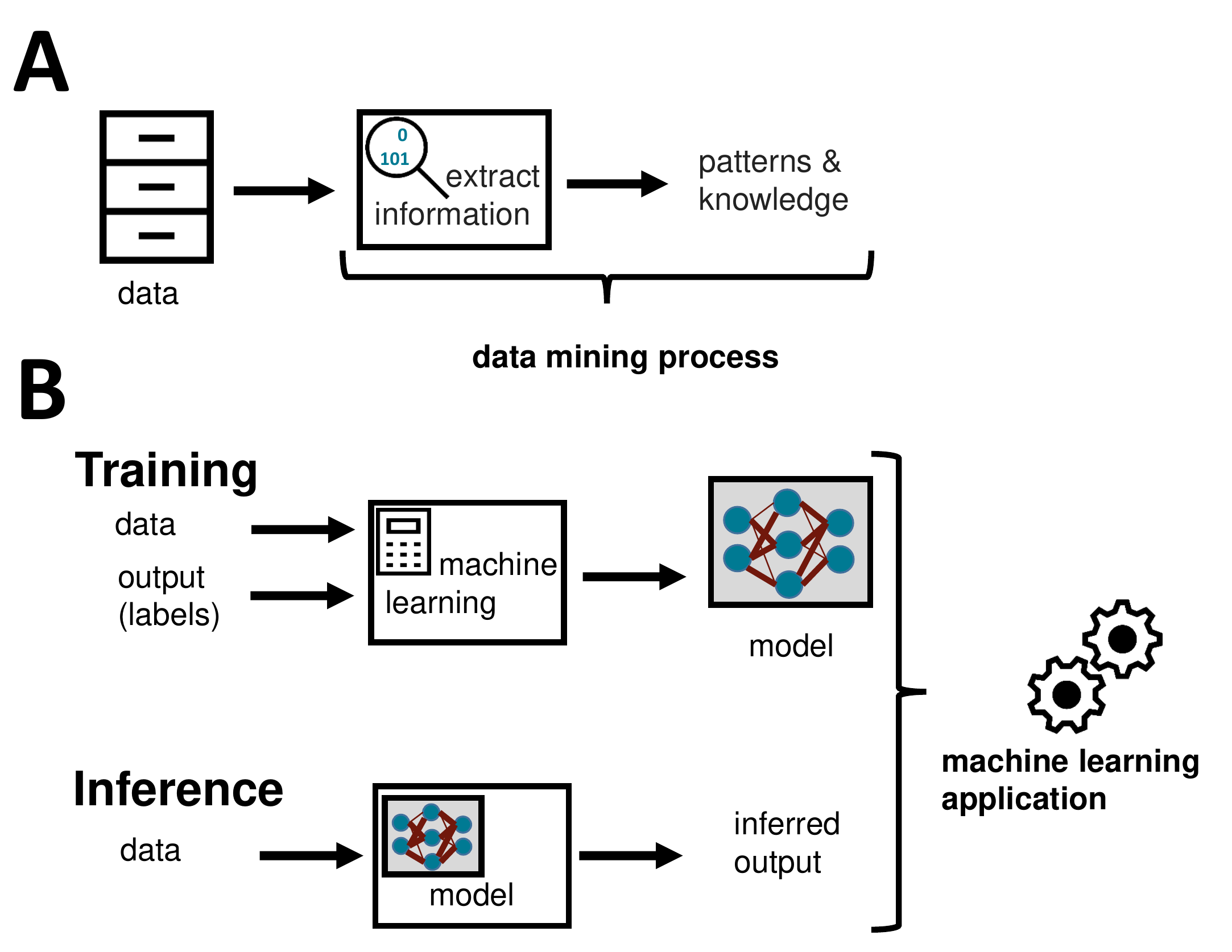}
	\caption{Difference between data mining processes and machine learning applications. A) In the data mining process information is directly extracted from data to find pattern und gain knowledge. B) A machine learning application consists of two steps. A machine learning model on data is trained and applied to perform inference on new data. Note that the model itself can be studied to gain insight within a knowledge discovery process.}
	\label{Fig:Example:ml-types_alt}
\end{figure}

However, we have identified two major shortcomings of CRISP-DM:

First, CRISP-DM focuses on data mining and does not cover the application scenario of ML models inferring real-time decisions over a long period of time (see \cref{Fig:Example:ml-types_alt}). 
%where a ML model is monitored and maintained as an application. 
The ML model has to be adaptable to a changing environment or the model's performance will degrade over time, such that, a permanent monitoring and maintaining of the ML model is required after the deployment.
%The input distribution can   
%and the changing production data will 
%change over time and, thus, will drift away from the training data distribution. 
%Otherwhise, this could lead to a degradation of the performance of the model 
%which is why the model has to be monitored and maintained. 

Second, and more worrying, CRISP-DM lacks guidance on quality assurance methodology. 

This oversight is particularly evident in comparison to standards in the area of information technology \cite{ieee10741997} but also apparent in alternative process models for data mining \cite{masemefe09, semma}. In our definition, quality is not only defined by the product's fitness for its purpose~\cite{Mariscal_2010}, but the quality of the task executions in any phase during the development of a ML application. This ensures that errors are caught as early as possible to minimize costs in the later stages during the development.
The initial effort and cost to perform the quality assurance methodology is expected to outbalance the risk of fixing errors in a later state, that are typically more expensive due to increased project complexity \cite{surange2015implementation, IBMMuthusamy}.
Our process model follows the principles of CRISP-DM, in particular by keeping the model industry and application neutral, but is modified to the particular requirements of ML applications and proposes quality assurance methodology that became industry best practice. Our contributions focus primarily on the technical tasks needed to produce evidence that every step in the development process is of sufficient quality to warrant the adoption into business processes. 

The following second section describes the related work and ongoing research in the development of process models for machine learning applications. In the third chapter, the tasks and quality assurance methodology are introduced for each process phase. Finally, a conclusion and an outlook are given in the fourth chapter.

\section{Related Work}
CRISP-DM defines a reference framework for carrying out data mining projects and sets out activities to be performed to complete a product or service. The activities are organized in six \emph{phases} (see \cref{tab:Process_model_differences}). The successful completion of a phase initiates the execution of the subsequent activity. CRISP-DM includes iterations of revisiting previous steps until success or completion criteria are met. It can be therefore characterized as a waterfall life cycle with backtracking~\cite{masemefe09}. During the development of applications, processes and tasks to be performed can be derived from the standardized process model. Methodology instantiates these tasks, i.e.\ stipulates how to do a task (or how it should be done).

For each activity, CRISP-DM defines a set of \emph{(generic) tasks} that are stable and general. Hereby, tasks are called \emph{stable} when they are designed to keep the process model up to date with new modeling techniques to come and \emph{general} when they are intended to cover many possible project scenarios. 
%We refer to~\cite{crisp-dm} for an exhaustive listing and description of tasks involved in data mining \cite{crisp-dm} . 
CRISP-DM has been specialized, e.g., to incorporate temporal data mining (CRISP-TDM; \cite{casmmctr09}), null-hypothesis driven confirmatory data mining (CRISP-DM0; \cite{hemc10}), evidence mining (CRISP-EM; \cite{vewawi07}), and data mining in the healthcare (CRISP-MED-DM; \cite{niaksu15}).

Complementary to CRISP-DM, process models for ML applications have been proposed \cite{Amershi_2019,breck2017ml}  (see \Cref{tab:Process_model_differences}). \citet{Amershi_2019} conducted an internal study at Microsoft on challenges of ML projects and derived a process model with nine different phases. However, their process model lacks quality assurance methodology and does not cover the business needs. \citet{breck2017ml} proposed 28 specific tests to quantify issues in the ML pipeline to reduce the technical debt \cite{sculley2015hidden} of ML applications. These tests estimate the production readiness of a ML application, i.e., the quality of the application in our context. However, their tests do not completely cover all project phases, e.g., excluding the business understanding activity. Practical experiences reveal that business understanding is a necessary first step that defines the success criteria and the feasibility for the subsequent tasks. Without considering the business needs, the ML objectives might be defined orthogonal to the business objectives and causes to spend a great deal of effort producing the rights answers to the wrong questions.

To our knowledge, \citet{masemefe09} were the first to consider quality in the context of process models for data mining. Borrowing ideas from software development, their work suggests creating traceability, test procedures, and test data for challenging the product's fitness for its purpose during the evaluation phase. 

We address these issues by devising a process model for the development of practical ML applications. 
% we don't have to state what we are not doing... <ah>
%This scope implicates, that our process model will not provide guidelines for the development of the best ML model. One cannot guarantee that modeling technique X is best suited for data set Y for all possible future cases because future development could result in better solutions. 
In addition, we will provide a curated list of references for an in-depth analysis on the specific tasks.

\section{Quality Assurance in Machine Learning Projects}
We propose a process model that we call \textbf{CR}oss-\textbf{I}ndustry \textbf{S}tandard \textbf{P}rocess model for the development of \textbf{M}achine \textbf{L}earning applications with \textbf{Q}uality assurance methodology (CRISP-ML(Q)) to highlight its compatibility to CRISP-DM. It is designed for the development of machine applications i.e.\ application scenarios where a ML model is deployed and maintained as part of a product or service (see \cref{Fig:Example:ml-types_alt}).  

\begin{figure}[htb]
\centering
\includegraphics[width=0.3\textwidth]{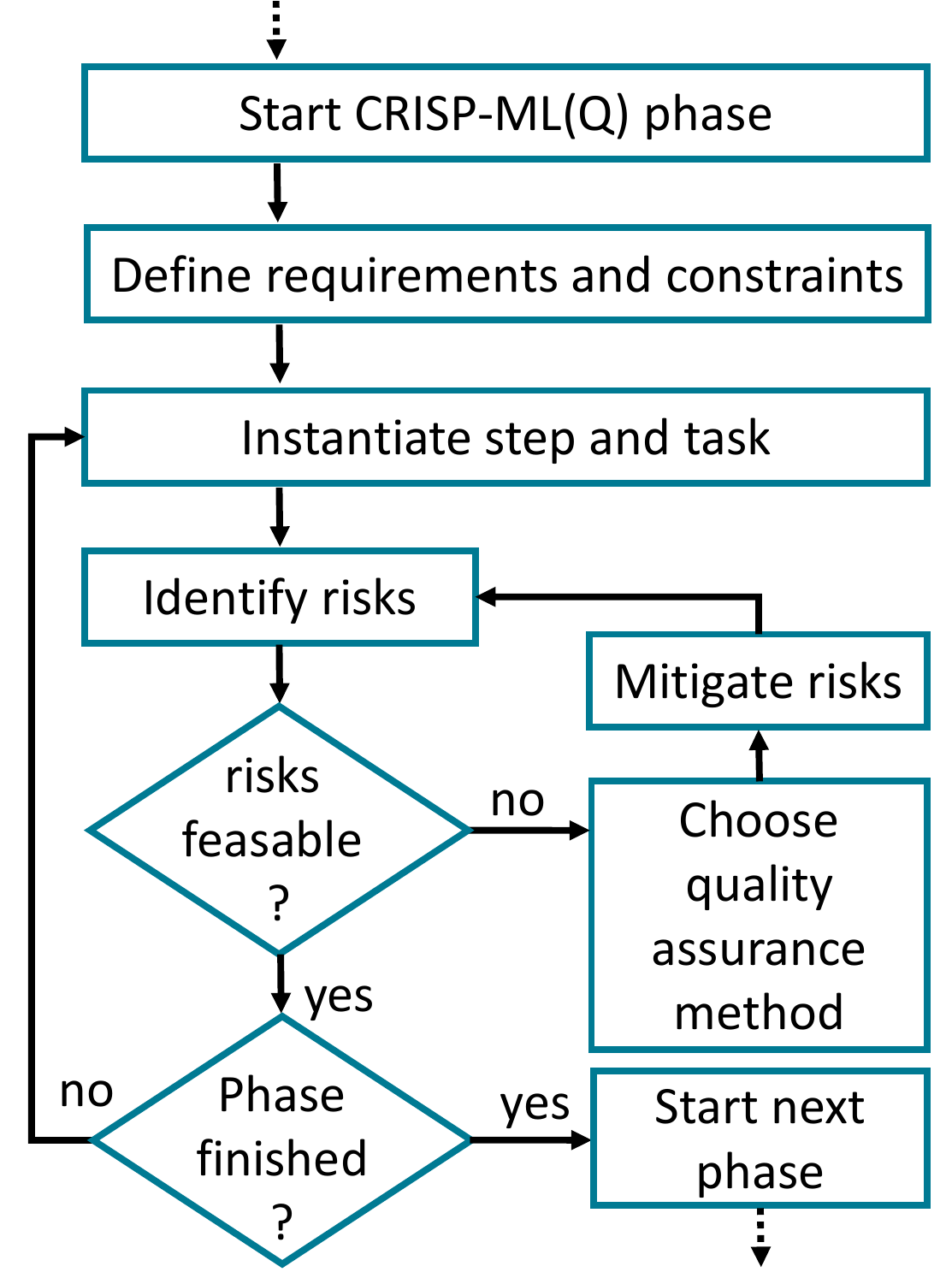}
\caption{Illustration of the CRISP-ML(Q) approach for quality assurance. The flow chart shows the instantiation of one specific tasks in a development phase, and the dedicated steps to identify and mitigate risks.}
\label{Fig:Example:flowchart}
\end{figure}

As a first contribution, \textit{quality assurance methodology} is introduced in each phase and task of the process model (see \cref{Fig:Example:flowchart}). The quality methodology serves to mitigate risks that affect the success and efficiency of the machine learning application. As a second contribution, CRISP-ML(Q) covers a \textit{monitoring and maintenance phase} to address risks of model degradation in a changing environment. This extends the scope of the process model as compared to CRISP-DM, see \Cref{tab:Process_model_differences}. 
Moreover, \textit{business and data understanding} are merged into a single phase because industry practice has taught us that these two activities, which are separate in CRISP-DM, are strongly intertwined, since business objectives can be derived or changed based on available data (see \Cref{tab:Process_model_differences}). A similar approach has been outlined in the W-Model \cite{wmodel}.  
%See \Cref{tab:Process_model_differences} for an overview on the six phases of CRISP-ML(Q).

\begin{table}[htb]
\scriptsize
\def\arraystretch{1.25}
\resizebox{\columnwidth}{!}{
\begin{tabular}{|p{1.9cm}|p{1.9cm}|p{1.6cm}|p{.8cm}|p{1cm}|}
\toprule 
 \multicolumn{1}{|c|}{\multirow{1}{*}{CRISP-ML(Q)}} & \multicolumn{1}{c|}{\multirow{1}{*}{CRISP-DM}} & \multicolumn{1}{c|}{\multirow{1}{*}{\citet{Amershi_2019}}} & \multicolumn{2}{c|}{\multirow{1}{*}{\citet{breck2017ml}}} \\[5pt] 
\toprule 
 \multirow{2}{*}{\makecell[l]{ \vspace{0.1cm}Business \& \\ \textcolor{DaimlerRed}{Data} \\ \textcolor{DaimlerRed}{Understanding} }} & Business Understanding & \textcolor{DaimlerPetrol}{Require\-ments} & \multicolumn{2}{c|}{- }\\ \cline{3-5} \cline{2-2} 
  & \textcolor{DaimlerRed}{Data Understanding}     &\textcolor{DaimlerRed}{Collection} &  \multirow{4}{*}{\makecell[l]{\textcolor{DaimlerRed}{Data} }} & \\ \cline{1-3}  %\cline{5-5}
 \multirow{3}{*}{\makecell[l]{ \textcolor{DaimlerRed} {Data} \\ \textcolor{DaimlerRed}{Preparation} }} & \multirow{3}{*}{\makecell[l]{ \textcolor{DaimlerRed}{Data} \\ \textcolor{DaimlerRed}{Preparation} }} & \textcolor{DaimlerRed}{Cleaning} & & \multirow{6}{*}{\makecell[l]{Infra- \\ struc\-ture}} \\ \cline{3-3}
 & &\textcolor{DaimlerRed}{Labeling} & & \\ \cline{3-3}
 & &  \textcolor{DaimlerRed}{Feature Engineering} & & \\ \cline{1-4}
 \textcolor{DaimlerPetrol}{Modeling}   &  \textcolor{DaimlerPetrol}{Modeling}  &\textcolor{DaimlerPetrol}{Training} & \textcolor{DaimlerPetrol}{Model} & \\ \cline{1-4}
 \textcolor{DaimlerPetrol}{Evaluation} & \textcolor{DaimlerPetrol}{Evaluation} & \textcolor{DaimlerPetrol}{Evaluation} & - & \\ \cline{1-4}
 \textcolor{DaimlerPetrol}{Deployment} &\textcolor{DaimlerPetrol}{Deployment} &\textcolor{DaimlerPetrol}{Deployment} & - & \\ \cline{1-5}
     \textcolor{DaimlerPetrol}{\makecell[l]{Monitoring \& \\ Maintenance }}        & - & \textcolor{DaimlerPetrol}{Monitoring} & \multicolumn{2}{c|}{\textcolor{DaimlerPetrol}{Monitoring}} \\ \hline
\end{tabular}
}
\caption{Comparing different process models for DM and ML projects. Business and data understanding phases are merged in CRISP-ML(Q) and a separate maintenance phase is introduced in comparison to CRISP-DM.
\citet{Amershi_2019} and \citet{breck2017ml} lack the business understanding phase. Deep red color highlight data and petrol blue color model related phases.}
\label{tab:Process_model_differences}
\end{table}

In what follows, we describe selected tasks from CRISP-ML(Q) % for developing ML applications 
and propose quality assurance methodology to determine whether these tasks were performed according to current standards from industry best practice and academic literature, which have proven to be general and stable and are suitable to mitigate the task specific risks. The selection reflects tasks and methods that we consider the most important.

The flow chart in \cref{Fig:Example:flowchart} explains the CRISP-ML(Q) approach for quality assurance. Requirements and constraints define the objectives of a generic phase, instantiate specific steps and tasks and identify risks, that can affect the efficiency and success of the ML application. If risks aren't feasible, appropriate quality assurance methods are chosen to mitigate risks in an iterative approach using %CRISP-ML(Q) as a 
guidelines and checklists. While general risk management has diverse disciplines \cite{ieee10741997}, this approach focuses on risks that affect the efficiency and success of the ML application and require technical tasks for risk mitigation.

%We cannot claim that the selection is complete, but it reflects tasks and methods that we consider the most important.
Note that the processes and quality measures in this document are not designed for safety-critical systems. Safety-critical systems might require different or additional processes and quality measures. 

\subsection{Business and Data Understanding} \label{subsec:bd-understanding} 
The initial phase is concerned with tasks to define the business objectives and translate it to ML objectives, to collect and verify the data quality and to finaly assess the project feasibility.
%and, finally, decide upon whether the project should be continued.

\subsubsection{Define the Scope of the ML Application} \label{subsec:scope} 
CRISP-DM names the data scientist responsible to define the scope of the project. However, in daily business, the separation of domain experts and data scientists carries the risk, that the application will not satisfy the business needs. Moreover, the availability of training samples will to a large extent influence the feasibility of the data-based application~\cite{Amershi_2019}. It is, therefore, best practice to merge the requirements of the business unit with ML requirements while keeping in mind data related constraints in a joint step. 
%This process might shift the scope, or even cause a delay or stop the development of the ML application.

\subsubsection{Success Criteria} \label{subsec:success-criteria} 
We propose to measure the success criteria of a ML project on three different levels: the business success criteria, the ML success criteria and the economic success criteria. 
According to the IEEE standard for developing software life cycle processes~\cite{ieee10741997}, the requirement \emph{measurable} is one of the essential principles of quality assurance methodology. 
In addition, each success criterion has to be defined in alignment to each other and with respect to the overall system requirements \cite{denso} to prevent contradictory objectives. 
%The success criteria will vary depending on the application considered.

\textit{Business Success Criteria:} Define the purpose and the success criteria of the ML application from a business point of view. %For example achieving a significant better application performance that outweights development costs.  
For example, if an ML application is planned for a quality check in production and is supposed to outperform the current manual failure rate of 3\%, the business success criterion could be derived as e.g.\ "failure rate less than 3\%".

\textit{ML Success Criteria:} Translate the business objective into ML success criteria (see \cref{tab:Properties of models}). It is advised to define a minimum acceptable level of performance to meet the business goals (e.g.\ for a Minimal Viable Product (MVP))
% and improved further on. 
In the mentioned example, the minimal success criterion is defined as "accuracy greater 97\%", but data scientists might optimize further, for example, the true-positive rate to miss items with quality issues.

\textit{Economic Success Criteria:} It is best practice to add an economic success criterion in the form of a Key Performance Indicator (KPI) to the project. A KPI is an economical measure for the relevance of the ML application. In the mentioned example, a KPI can be defined as "cost savings with automated quality check per part".

\subsubsection{Feasibility} \label{subsubsec:Assess Situation}
Checking the feasibility before setting up the project is considered best practice for the overall success of the ML approach \cite{watanabe2019preliminary} and can minimize the risk of premature failures due to false expectations.
A feasibility test of the ML application should assess the situation and whether further development should be pursued. 
It is crucial, that the assessment includes the availability, size and quality of the training sample set.
In practice, a major source of project delays is the lack of data availability (see \cref{data_collection}). 
A small sample size carries the risk of low performance on out-of-sample data~\cite{Rudin_2019_secrets}. The risk might be mitigated by e.g.\ adding domain knowledge or increasing data quality. However, if the sample size is not sufficient the ML project should be terminated or put on hold at this stage.  
%Further assessments include legal constraints, the applicability of the ML technology and preliminary works. 

\textit{Applicability of ML technology:} Literature search for either similar applications on a similar domain or similar methodological approaches on a different domain could assess the applicability of the ML technology.
It is common to demonstrate the feasibility of a ML application with a proof of concept (PoC) when the ML algorithm is used for the first time in a specific domain. 
%However, if an ML application has been used before successfully, the development of a PoC could amount to a loss in time and can be skipped. 
If a PoC already exists, setting up a software project that focuses on the deployment directly is more efficient, e.g.\ in case of yet another price estimation of used cars~\cite{pudaruth2014predicting}. 
%An example from the automotive industry is the price estimation of used cars using ML models \cite{pudaruth2014predicting}. ML models are state-of-the-art in any car vending platform and, therefore, doesn't require a proof of concept. 

\textit{Legal constraints:} It is beyond the scope of this paper to discuss legal issues but they are essential for any business application \cite{reed2016responsibility, bibal2020legal}. 
% but it is essential to include the legal department to check for legal constraints as those could impede the feasibility of the project. %Legal constraints could be, for example, defined by the licenses on the used software or data, the necessary data anonymization or safety requirements. 
Legal constraints are frequently augmented by ethical and social considerations like fairness and trust \cite{Binns_2018,Corbett-Davies_2018,Arrieta_2019}. 

\textit{Requirements on the application:} The success criteria that have been defined in \cref{subsec:success-criteria} have to be augmented with requirements that arise from running the application in the target domain or if not accessible an assumed target domain \cite{denso}. The requirements include  robustness, scalability, explainability and resource demand and are used for the development and verification in later phases (see \cref{subsec:modeling}). The challenge during the development is to optimize the success criteria while not violating the requirements and constraints. 
%Requirements could be, for example, the inference time of a prediction, the memory size of the model (considering it has to be deployed on hardware with limited memory), the performance and the robustness of a model or on the quality of the data . 
%\textcolor{red}{not clear how the Requirements on the application are derived from e.g. the Ml success criteria..Binh: Its the other way around. You derive the ML success criteria from the requirements on the application}
%The challenge during the development is to optimize the success criteria while not violating the requirements and constraints. 

\subsubsection{Data Collection}\label{data_collection}
Costs and time is needed to collect a sufficient amount of consistent data by preparing and merging data from different sources and different formats (see \cref{phase:data_preparation}). 
%Before starting to collect data, estimate roughly which and how many data is necessary and what costs occur. Data could be collected from many different sources and have to be merged into one data set. Different data sets could have different formats, features or labels which has to be considered during the merge. However, in the case if there is no data available or very few data, it might be necessary to create an infrastructure to collect the data. 
%The recording of additional data could be done using, for example, techniques like active learning \cite{cohn1996active} or Bayesian optimization \cite{osborne2009gaussian}. 
A ML project might be delayed until the data is collected or could even be stopped if the collection of data of sufficient quality (see \cref{subsubsec:Data Quality Verification}) is not feasible.
%could act as an exit criteria if the collection of new data not feasible.

\textit{Data version control:} Collecting data is not a static task but rather an iterative task. Modification on the data set (see \cref{phase:data_preparation})
%by adding and removing data, modifications on the selected features or labels 
should be documented to mitigate the risk of obtaining irreproducible or wrong results. 
Version control on the data is one of the essential tools to assure reproducibility and quality as it allows to track errors and unfavorable modifications during the development.

\subsubsection{Data Quality Verification}\label{subsubsec:Data Quality Verification}
The following three tasks examine whether the business and ML objectives can be achieved with the given quality of the available data. 
A ML project is doomed to fail if the data quality is poor.
%ML models depend heavily on the training data and, as a consequence, poor data often leads to poor models. 
The lack of a certain data quality will trigger the previous data collection task (see \cref{data_collection}). 
%The data quality verification includes three tasks:  describe the data, define requirements on the data and verify the data.

\textit{Data description:} The data description forms the basis for the data quality verification. 
A description and an exploration of the data is performed to gain insight about the underlying data generation process. 
The data should be described on a meta-level %e.g.\ a pedestrian should have two legs and two arms 
and by their statistical properties. 
Furthermore, a technically well funded visualization of the data should help to understand the data generating process \cite{mcqueen2016megaman}. 
Information about format, units and description of the input signals is expanded by domain knowledge. 

\textit{Data requirements:} The data requirements can be defined either on the meta-level or directly in the data and should state the expected conditions of the data, i.e. whether a certain sample is plausible. 
The requirements can be, e.g., the expected feature values (a range for continuous features or a list for discrete features), the format of the data and the maximum number of missing values. 
The bounds of the requirements has to be defined carefully 
%by the development team 
to include all possible real world values but discard non-plausible data. 
Data that does not satisfy the expected conditions could be treated as anomalies and have to be evaluated manually or excluded automatically. To mitigate the risk of anchoring bias in the definition phase discussing the requirements with a domain expert is advised~\cite{breck2017ml}. Documentation of the data requirements could be done in the form of a schema~\cite{poly17,schelter2019unit}.

\textit{Data verification:} The initial data, added data but also the production data has to be checked according to the requirements (see \cref{subsec:Monitoring and Maintenance}). In cases where the requirements are not met, the data will be discarded and stored for further manual analysis. This helps to reduce the risk of decreasing the performance of the ML application through adding low-quality data and helps to detect varying data distributions or unstable inputs.
% e.g. the units of one of the features changed from kilograms to grams during an update. 
%Finally, check the coverage of the data by plotting histograms and computing the statistics of the data to assure a sufficient representation of extreme cases.
To mitigate the risk of insufficient representation of extreme cases, it is best practice to use data exploration techniques to investigate the sample distribution.

\subsubsection{Review of Output Documents}
The Business \& Data Understanding phase delivers the scope for the development (\cref{subsubsec:Assess Situation}), the success criteria (\cref{subsec:success-criteria}) of a ML application and a data quality verification report (\cref{subsubsec:Data Quality Verification}) to approve the feasibility of the project. The output documents need to be reviewed to rank the risks and define the next tasks. If certain quality criteria are not met, re-iterations of previous tasks are possible.

\subsection{Data Preparation}\label{phase:data_preparation}
Building on the experience from the preceding data understanding phase, data preparation serves the purpose of producing a data set for the subsequent modeling phase. However, data preparation is not a static phase and backtracking circles from later phases are necessary if, for example, the modeling phase or the deployment phase reveal erroneous data. To path the way towards ML lifecycle in a later phase, methods for data preparation that are suitable for automation as demonstrated by \citet{make3010004} are preferable.

\subsubsection{Select Data}\label{subsec:sel-data}
%Select data is the task of selecting a relevant subset of representative data and features for the training, validation and test set.

\textit{Feature selection:} Selecting a good data representation based on the available measurements is one of the challenges to assure the quality of the ML application. It is best practice to discard underutilized features as they provide little to none modeling benefit but offer possible loopholes for errors i.e. instability of the feature during the operation of the ML application \cite{sculley2015hidden}. In addition, the more features are selected the more samples are necessary. Intuitively an exponentially increasing number of samples for an increasing number of features is required to prevent the data from becoming sparse in the feature space. This is termed as the curse of dimensionality~\cite{Keogh_2017,Bishop_2007}. Thus, it is best practice to select just necessary features. A checklist for the feature selection task is given in \cite{Guyon:2003:IVF:944919.944968}. Note that data often forms a manifold of lower dimensions in the feature space and models have to learn this respectively \cite{braun2008relevant}.  Feature selection methods can be separated into three categories: 1) \textit{filter methods} select features from data without considering the model, 2) \textit{wrapper methods} use a learning model to evaluate the significance of the features and 3) \textit{embedded methods} combines the feature selection and the classifier construction steps. A detailed explanation and in-depth analysis on the feature selection problem are given in \cite{hira2015review, Saeys:2007:RFS:1349154.1349169, Chandrashekar:2014:SFS:2577586.2577699, Guyon:2006:FEF:1208773}.  Feature selection could carry the risk of selection bias but could be reduced when the feature selection is performed within the cross-validation of the model (see \cref{subsec:modeling}) to account for all possible combinations  \cite{ambroise2002selection}.

However, the selection of the features should not be relied purely on the validation and test error but should be analyzed by a domain expert as potential biases might occur due to spurious correlation in the data. Lapuschkin \textit{et al.} \cite{Lapuschkin_FisherVectorLRP,lapuschkin2019unmasking} showed that classifiers could exploit spurious correlations, here the copyright tag on the horse class, to obtain a remarkable test performance and, thus, fakes a false sense of generalization. In such cases, explanation methods \cite{samek2019explainable} could be used to highlight the significance of features (see \cref{subsec:Evaluation}) and analyzed from a human's perspective.

\textit{Data selection:} Discarding samples should be well documented and strictly based on objective quality criteria. However, certain samples might not satisfy the necessary quality i.e. doesn't satisfy the requirements defined in \cref{subsubsec:Data Quality Verification} and are not plausible and, thus, should be removed from the data set. 
%ML models rest upon the assumption of an adequate number of samples and, therefore, the predictive performance of the model increases by adding more samples \cite{Vapnik_1995, simard}. 

\textit{Unbalanced Classes:} In cases of unbalanced classes, where the number of samples per class is skewed, different sampling strategies could improve the results. Over-sampling of the minority class and/or under-sampling of the majority class \cite{lawrence1998neural, Chawla_2002, Batista_2004, Lemaitre:2017:IPT:3122009.3122026} have been used. Over-sampling increases the importance of the minority class but could result in overfitting on the minority class. Under-Sampling by removing data points from the majority class has to be done carefully to keep the characteristics of the data and reduce the chance of introducing biases. However, removing points close to the decision boundary or multiple data points from the same cluster should be avoided. Comparing the results of different sampling techniques' reduces the risk of introducing bias to the model. 

\subsubsection{Clean Data}
%Cleaning data addresses the noise in the data and the imputation of missing values. 
%If a feature or sample subsets cannot be sufficiently cleaned it might be better to discard these data, i.e. returning to the data selection task described before.

\textit{Noise reduction:} The gathered data often includes, besides the predictive signal, noise and unwanted signals from other sources. Signal processing filters could be used to remove the irrelevant signals from the data and improve the signal-to-noise ratio \cite{walker2002primer, Lyons:2004:UDS:993484}. However, filtering the data should be documented and evaluated because of the risk that an erroneous filter could remove important parts of the signal in the data. 

\textit{Data imputation:} To get a complete data set, missing, NAN and special values could be imputed with a model readable value. Depending on the data and ML task the values are imputed by mean or median values, interpolated, replaced by a special value symbol \cite{che2018recurrent} (as the pattern of the values could be informative), substituted by model predictions \cite{biessmann2018deep}, matrix factorization \cite{Koren:2009:MFT:1608565.1608614} or multiple imputations \cite{murray2018multiple, white2011multiple, azur2011multiple} or imputed based on a convex optimization problem \cite{JMLR:v18:17-073}. To reduce the risk of introducing substitution artifacts, the performance of the model should be compared between different imputation techniques. 

\subsubsection{Construct Data}\label{subsec:construct-data}
%Constructing data includes the tasks of deriving new features (feature engineering) and constructing new samples (data augmentation).

\textit{Feature engineering:} New features could be derived from existing ones based on domain knowledge. This could be, for example, the transformation of the features from the time domain into the frequency domain, discretization of continuous features into bins or augmenting the features with additional features based on the existing ones.
% e.g. squaring, taking the square root, the log, the inverse, etc. 
In addition, there are several generic feature construction methods, such as clustering \cite{coates2012learning}, dimensional reduction methods such as Kernel-PCA \cite{scholkopf1997kernel} or auto-encoders \cite{rumelhart1985learning}.
%This could aid the learning process and improves the predictive performance of the model. 
Nominal features and labels should be transformed into a one-hot encoding while ordinal features and labels are transformed into numerical values. However, the engineered features should be compared against a baseline to assess the utility of the feature. Underutilized features should be removed.
Models that construct the feature representation as part of the learning process, e.g. neural networks, avoid the feature engineering steps \cite{Goodfellow_2016}.
% unless prior knowledge is available. 

\textit{Data augmentation:} Data augmentation utilizes known invariances in the data to perform a label preserving transformation to construct new data. The transformations could either be performed in the feature space \cite{Chawla_2002} or input space, such as applying rotation, elastic deformation or Gaussian noise to an image \cite{wong2016understanding}. Data could also be augmented on a meta-level, such as switching the scenery from a sunny day to a rainy day. This expands the data set with additional samples and allows the model to capture those invariances. 
%It is recommended to perform data augmentation in the input space if invariant transformations are known \cite{wong2016understanding}.
  
\subsubsection{Standardize Data}
%The data and the format of the data should be standardized to get a consistent data set i.e. transforming into a common file format, normalization of the features and labels, the usage of common units and standards.

\textit{File format:} Some ML tools require specific variable or input types (data syntax). Indeed in practice, the comma separated values (CSV) format is the most generic standard (RFC 4180).
%, it has been proven as a method for PoC studies or to obtain an early MVP. 
ISO 8000 recommends the use of SI units according to the International System of Quantities. Defining a fix set of standards and units, helps to avoid the risks of errors in the merging process and further in detecting erroneous data (see \cref{subsubsec:Data Quality Verification}).
%i.e. doesn't satisfy the requirements made in \cref{subsubsec:Data Quality Verification}.

\textit{Normalization:} Without proper normalization, the features could be defined on different scales and might lead to strong bias to features on larger scales. In addition, normalized features lead to faster convergence rates in neural networks than without \cite{lecun2012efficient, DBLP:journals/corr/IoffeS15}. Note that the normalization, applied to the training set has to be applied also to the test set using the same normalization parameters.

\subsection{Modeling}\label{subsec:modeling}
\begin{figure}[!h]
	\centering
	\includegraphics[width=0.45\textwidth]{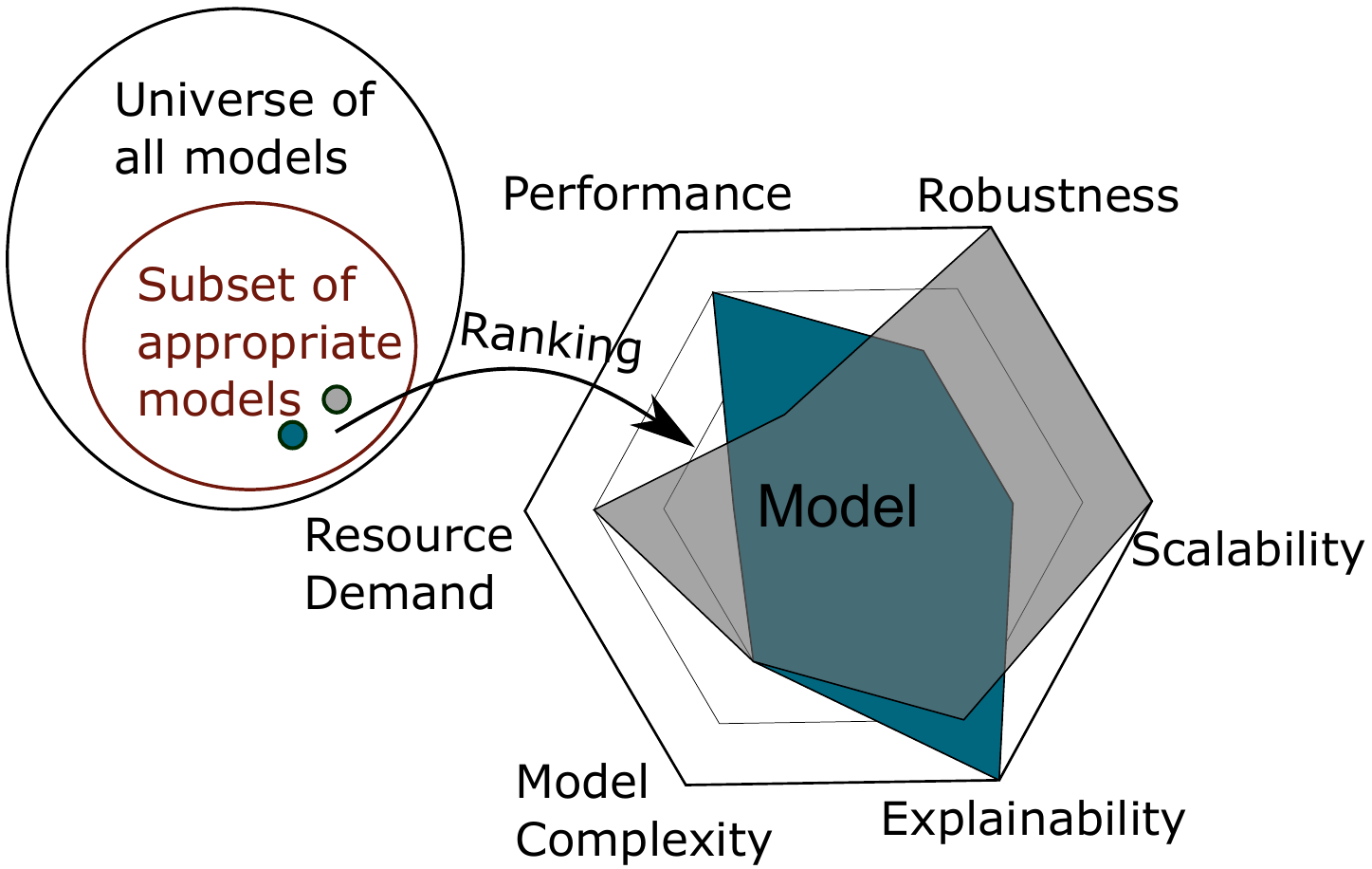}
	\caption{An outline of the modeling phase. Only a subset of models fulfill the constraints and requirements defined in \cref{subsec:bd-understanding} and \cref{subsubsec:Assess Situation} and have to be evaluated using quality measures (see \cref{tab:Properties of models})}.
	\label{Fig:Modeling process}
\end{figure}

The choice of modeling techniques depends on the ML and the business objectives, the data and the boundary conditions of the project the ML application is contributing to. The requirements and constraints that have been defined in \cref{subsec:bd-understanding} are used as inputs to guide the model selection to a subset of appropriate models. The goal of the modeling phase is to craft one or multiple models that satisfy the given constraints and requirements. An outline of the modeling phase is depicted in \cref{Fig:Modeling process}.

\textit{Literature research on similar problems:} It is best practice to screen the literature (e.g.\ publications, patents, internal reports) for a comprehensive overview on similar ML tasks, since ML has become an established tool for a wide number of applications. New models can be based on published insights and previous results can serve as performance baselines.

\textit{Define quality measures of the model:} The modeling strategy has to have multiple objectives in mind \cite{modiTFX}.
%\cite{modiTFX} suggest to evaluate the model by two properties: a model has to be safe to serve and has to have the desired prediction quality. 
We suggest to evaluate the models on at least six complementary properties (see \cref{Fig:Modeling process}). Besides a \emph{performance metric}, soft measures such as \emph{robustness}, \emph{explainability}, \emph{scalability}, \emph{resource demand} and \emph{model complexity} have to be evaluated (see \cref{tab:Properties of models}). The measures can be weighted differently depending on the application. In practical application, explainability~\cite{Rudin_2019,Rudin_2019_secrets,schmidt2019quantifying} or robustness might be valued more than accuracy. Additionally, the model's \emph{fairness}~\cite{Binns_2018,Corbett-Davies_2018} or \emph{trust} might have to be assessed and mitigated. 
%In a case study, \cite{schmidt2019quantifying} showed empirically that highlighting the three most important features of a ML model could help to improve the performance of a human in text classification problems. 

\begin{table}[!h]
\scriptsize
\begin{center}
\bgroup
\def\arraystretch{1.4}%
\arrayrulecolor{white}
\arrayrulewidth=0.6mm
\resizebox{\columnwidth}{!}{
  \begin{tabular}{ m{1.6cm}!{\color{white}\VRule[0.6mm]} m{6.4cm}}
    \hline \rowcolor{tableLineOne}
			Performance & The model's performance on unseen data.\\ \hline \rowcolor{tableLineTwo}
			Robustness & The model's resiliency to inconsistent inputs and to failures in the execution environment.  \\ \hline \rowcolor{tableLineOne}	
			Scalability & The model's ability to scale to high data volume in the production system.  \\ \hline \rowcolor{tableLineTwo}
			Explainability & The model's direct or post-hoc explainability. \\ \hline \rowcolor{tableLineOne}
			Model Complexity & The model's capacity should suit the data complexity. \\ \hline \rowcolor{tableLineTwo}
			Resource Demand & The model's resource demand for deployment. \\ \hline \rowcolor{tableLineOne}
			%\textcolor{red}{Fairness} & \textcolor{red}{The model's decisions are equitable.} \\ \hline
  \end{tabular}
}
  \egroup
	\caption{Quality measure of machine learning models}
	\label{tab:Properties of models}
\end{center}
\end{table}

%% ToDo: Check that examples etc.. are in the text...
%			Robustness & Resiliency of the model to inconsistent inputs e.g. adversarial attacks, out-of-distribution samples, anomalies and distribution shifts and to failures in the underlying execution environment e.g. sensor, actuators and computational platform.  \\ \hline \rowcolor{tableLineOne}	
%			Scalability & The property of the model to scale to high data volume during the training and re-training in the production system. Complexity analysis on the execution time and hardware demand dependent on the number of samples and feature dimension. \\ \hline \rowcolor{tableLineTwo}
%			Explainability & Models could be either directly explainable or given by post-hoc explanations. The decisions of explainable models could be inspected manually and could increase the user acceptance. In addition, uncertainty and confidence estimates provide guidance on indecisive decisions. \\ \hline \rowcolor{tableLineOne}
%			Model Complexity & Models with large capacities overfit easily on small data sets. Assure that the capacity of your model suits the complexity of your data and use proper regularization. \\ \hline \rowcolor{tableLineTwo}
%			Resource Demand & The model has to be deployed on hardware and is restricted by its memory. In addition, the inference time has to be considered dependent on the application.\\ \hline

\textit{Model Selection:} There are plenty of ML models and introductory books on classical methods \cite{Bishop_2007, scholkopf2002learning} and Deep Learning \cite{Goodfellow_2016} can be used to compare and understand their characteristics. The model selection depends on the data and has to be tailored to the problem. There is no such model that performs the best on all problem classes (\textit{No Free Lunch Theorem for ML} \cite{Wolpert:1996:LPD:1362127.1362128}). It is best practice to start with models of lower capacity, which can serve as baseline, and gradually increase the capacity. Validating each step assures its benefit and avoid unnecessary complexity of the model.

\textit{Incorporate domain knowledge:} In practice, a specialized model for a specific task performs better than a general model for all possible tasks. However, adapting the model to a specific problem involves the risk of incorporating false assumption and could reduce the solution space to a non-optimal subset. Therefore, it is best practice to validate the incorporated domain knowledge in isolation against a baseline. Adding domain knowledge should always increase the quality of the model, otherwise, it should be removed to avoid false bias. 
%
% Zien et al.\ \cite{Zien_2000} showed that specialized kernels could improve the performance of the model in recognizing translation initiation sites from nucleotide sequences. Another example are convolutional layers in neural networks are used because of the assumption that pixels in an image are locally correlated but also that the features are translation invariant. 

\textit{Model training:} The trained model depends on the learning problem and as such are tightly coupled. The learning problem contains an \textit{objective}, \textit{optimizer}, \textit{regularization} and \textit{cross-validation} \cite{Bishop_2007, Goodfellow_2016}. 
%An extensive and more formal description can be found in \cite{Bishop_2007, Goodfellow_2016}. 
The \textit{objective} of the learning problem depends on the application. Different applications value different aspects and have to be tweaked in alignment with the business success criteria. The objective is a proxy to evaluate the performance of the model. The \textit{optimizer} defines the learning strategy and how to adapt the parameters of the model to improve the objective. \textit{Regularization} which can be incorporated in the objective, optimizer and in the model itself is needed to reduce the risk of overfitting and can help to find unique solutions. \textit{Cross-validation} is performed for feature selection, to optimize the hyper-parameters of the model and to test its generalization property to unseen data ~\cite{muller2001introduction}. Cross-validation \cite{Bishop_2007} is based on a splitting of historical data in training, validation and testing data, where the latter is used as a proxy for the target environment \cite{DBLP:journals/corr/abs-1906-10742}.
%The data set is split into a training, a validation and a test set. While the training set is used in the learning procedure the validation set is used to test the generalization property of the model on unseen data and to tune the hyper-parameters \cite{muller2001introduction}. 
%The test set is used to estimate the generalization property of the model, see \cref{subsec:Evaluation}. 
%Hyper-parameters of all the models including the baselines should be optimized to validate the performance of the best possible model. 
Frameworks such as Auto-ML \cite{Hutter_2019, Feurer_2015} or Neural Architecture Search \cite{zoph2016neural} enables to partly automatize the hyper-parameters optimization and the architecture search.% but should be used with care.

\textit{Using unlabeled data and pre-trained models:} Labeling data could be very expensive and might limit the available data set size. Unlabeled data might be exploited in the training process, e.g.\ by performing unsupervised pre-training \cite{Erhan:2010:WUP:1756006.1756025, Dreher_2020} and semi-supervised learning algorithms \cite{NIPS2014_5352, Chapelle:2010:SL:1841234}. Complementary, \emph{transfer learning} could be used to pre-train the network on a proxy data set (e.g. from simulations) that resembles the original data to extract common features \cite{NIPS2014_5347}. %The proxy data can be obtained from simulations or closely related data sets. 
%Gathering simulated data is much cheaper and enables the construction of rare data points. For example, in industrial applications CAD models for all parts of a technical product are usually available and might be used for pre-training networks for object recognition and localization \cite{Andulkar_2018}. 

\textit{Model Compression:} Compression or pruning methods could be used to obtain a more compact model. In kernel methods low rank approximations of the kernel matrix is an essential tool to tackle large scale learning problems \cite{NIPS2000_1866, Drineas:2005:NMA:1046920.1194916}. Neural Networks use a different approach \cite{DBLP:journals/corr/abs-1710-09282} by either pruning the network weights \cite{frankle2018the} or applying a compression scheme on the network weights \cite{DBLP:journals/corr/abs-1907-11900}. 
%\cite{frankle2018the} was able to prune up to 90\% of the neural network weights while \cite{DBLP:journals/corr/abs-1907-11900} was able to compress the VGG16 ImageNet model by 63.6 times with no loss in accuracy. 
%A survey on neural network compression can be found in \cite{DBLP:journals/corr/abs-1710-09282}.

\textit{Ensemble methods:} Ensemble methods train multiple models to perform the decision based on the aggregate decisions of the individual models. The models could be of different types or multiple instantiations of one type. This results in a more fault-tolerant system as the error of one model could be absorbed by the other models. Boosting, Bagging or Mixture of Experts are mature techniques to aggregate the decision of multiple models \cite{rokach2010ensemble, zhou2002ensembling, opitz1999popular}. In addition, ensemble models are used to compute uncertainty estimates and can highlight areas of low confidence \cite{lakshminarayanan2017simple, gal2016dropout}.

\subsubsection{Assure reproducibility}
A major principle of scientific methods and the characteristics of robust ML applications is reproducibility. 
%A quality assurance method that is common to software engineering and science is to validate any result by peer-review. For instance, experiments can be validated by re-implementing the algorithms or running the given source code to reproduce the results. Ultimately, reproducibility is necessary to locate and debug errors.
However,  ML models are difficult to reproduce due to the mostly non-convex and stochastic training procedures and randomized data splits. 
%This has been addressed at the Neural Information Processing Systems (NeurIPS) 2019 with the creation of a Reproducibility Chair and a reproducibility checklist \cite{Pineau_2019}. 
It has been proposed to distinguish reproducibility on two different levels. First, one has to assure that the method itself is reproducible and secondly its results \cite{Pineau_2019}.
%To assuring the reproducibility an approach at two different levels, reproducibility of the method and of the results, as been proposed \cite{Pineau_2019}.  

\textit{Method reproducibility:} This task aims at reproducing the model from an extensive description or sharing of the used algorithm, data set, hyper-parameters and runtime environment (e.g.\ software versions, hardware and random seeds \cite{tatman2018practical}). 
The algorithm should be described in detail i.e.\ with (pseudo) code and on the meta-level including the assumptions. 
%The description should contain the version of the data sets used to train, validate and test the model (see \cref{subsubsec:Data Quality Verification}), a description of the modeling techniques, the chosen hyper-parameters, the software and its version being used to apply these techniques, the hardware it is been executed on and the random seeds \cite{Pineau_2019}.
%Additionally, \cite{tatman2018practical} proposed to provide an environment to run the code to avoid the \emph{it runs on my computer} problem. 

\textit{Result reproducibility:} It is best practice to validate the mean performance and assess the variance of the model on different random seeds \cite{henderson2018deep, sculley2018winner's}. Reporting only the top performance of the model \cite{pmlr-v97-bouthillier19a, henderson2018deep} is common but dubious practice. Large performance variances indicate the sensitivity of the algorithm and question the robustness of the model.
%it is questionable if the model could retain the performance after multiple updates
%It is common dubious practice to train multiple models with different random seeds and report the top performance of the model \cite{pmlr-v97-bouthillier19a, henderson2018deep}. This is deeply flawed as the variance of the performance is completely ignored and the result could be obtained by chance. Large variances dependent on the random seeds indicate the sensitivity of the algorithm and it is questionable if the model could retain the performance after multiple updates. It is, therefore, 

\textit{Experimental Documentation:} 
%As the modeling phase could cover many models and modifications in the data set, it is hard to keep track of all the changes, especially beneficial or unfavorable changes. 
Keeping track of the changed model's performance and its causes by precedent model modifications allows model comprehension by addressing which modifications were beneficial and improve the overall model quality.
% and which ones were harmful.  
The documentation should contain the listed properties in the \textit{method reproducibility} task.
% and can be used for debugging the code.
%Plan a documentation strategy and list the properties that should be documented. 
%For example, \cite{Vartak} showed a 
Tool-based approach on version control and meta-data handling while experimenting on ML models and hyper-parameters exist \cite{Vartak}.

\subsection{Evaluation} \label{subsec:Evaluation}
% Sort the introduction somehow to the specific tasks..
% %  somewhere: words of wisdoms: We outlined that quality oriented methodology is needed throughout the entire developing lifecycle (see Table 1), hence e.g. addressing overfitting in the evaluation phase is necessary but not sufficient.

% Validate: A major risk is caused by the fact that a complete test coverage of all possible inputs is not tractable because of the large input dimensions. 
%   However, extensive testing reduces the risk of failures. When testing, one has to always keep in mind that the stochastic nature of the data resulting in label noises bounds the test accuracy from the top. 
%   That means, 100\% test accuracy can be rarely achieved. 
%   One should keep in mind that testing the generalization error of the model with a train- validation- and test-split is insufficient when the data is sampled or collected by erroneous processes. 
\textit{Validate performance:} A risk occurs when information from a test set leak into the validation or even training set. % of the performance by using feedback signals from the test set to optimize the model. 
Hence, it is best practice to hold back an additional test set, which is disjoint from the validation and training set, stored only for a final evaluation and never shipped to any partner to be able to measure the performance metrics (blind-test). 
The test set should be assembled and curated with caution and ideally by a team of experts that are capable to analyze the correctness and ability to represent real cases. 
In general, the test set should cover the whole input distribution and consider all invariances, e.g.\ transformations of the input that do not change the label, in the data. %Invariances are transformations of the input that should not change the label of the data. 
Another major risk is that the test set cannot cover all possible inputs due to the large input dimensionality or rare corner cases, i.e.\ inputs with low probability of occuring~\cite{Zhou:2019:MTD:3314328.3241979, Tian:2018:DAT:3180155.3180220, Pei:2017:DAW:3132747.3132785}. Extensive testing reduces this risk \cite{DBLP:journals/corr/abs-1906-10742}.
%\cite{Zhou:2019:MTD:3314328.3241979, Tian:2018:DAT:3180155.3180220, Pei:2017:DAW:3132747.3132785} have shown that a highly sophisticated model for autonomous driving could not capture those invariances and found extreme cases which led to false predictions by transforming a picture taken on a sunny day to a rainy day picture or by darkening the picture. 
It is recommended to separate the teams and the procedures collecting the training and the test data to erase dependencies and avoid methodology dependence. %propagating %from the training set to the test set. 
%On that test set, the prior defined performance metrics should then be evaluated. 
Additionally, it is recommended to perform a sliced performance analysis to highlight weak performance on certain classes or time slices. 
%A full test set evaluation may mask flaws on certain slices.

% Robustness: When evaluating a ML solution to a business problem it is important to assure the correctness of the results but also to study its behavior on false inputs. 
\textit{Determine robustness:} A major risk occurs if ML applications are not robust to perturbed, e.g.\ noisy or wrong, or even designed adversarial input data as show by \citet{make1010011}.
This requires methods to statistically estimate the model's local and global robustness. One approach is adding different kinds of noisy or falsified input to the data or varying the hyper-parameters to characterize the model's generalization ability. 
%\cite{denso} show formal verification approaches, while \cite{DBLP:journals/corr/abs-1906-10742} validate robustness using cross-validation techniques on historical data.
Formal verification approaches \cite{denso} and robustness validation methods using cross-validation techniques on historical data \cite{DBLP:journals/corr/abs-1906-10742} exist.

%The robustness of the model, in terms of the model's ability to generalize to a perturbation of the data set, can be determined with K-fold cross-validation. Hereby, the algorithm is repeatedly validated by holding disjoint subsets of the data out of the training data as validation data. The mean performance and variance of the cross-validation can be analyzed to check the generalization ability of the model on different data sets. It might be beneficial to accept a lower training performance which can generalize well to unseen data than having a model that exhibits the inverse behavior. 
%Moreover, robustness should be checked when adding different kinds of noisy or wrong (e.g.\ missing values, NaNs, out of distribution data, error codes, etc.) input to the data or varying the hyper-parameters which characterize the model indirectly (e.g.\ the number of neurons in a deep neural network). 
%In addition, it is recommended to assure robustness of a model when given wrong inputs e.g. missing values, NaNs or data out of distribution as well as signals which might occur in case of malfunctions of input devices such as sensors. 
%A different challenge is given by adversarial examples \cite{goodfellow2014explaining} that perturbs the image by an imperceptible amount and fool classifiers to make wrong predictions. 
%A survey of current testing methods can be found in \cite{DBLP:journals/corr/abs-1906-10742}. 
The model's robustness should match the quality claims made in \cref{tab:Properties of models}.

\textit{Increase explainability for ML practitioner \& end user:} Explainability of a model helps to find errors and allows strategies, e.g.\ by enriching the data set, to improve the overall performance~\cite{Chakarov_2016}. In practice, inherently interpretable models are not necessary inferior to complex models in case of structured input data with meaningful features~\cite{Rudin_2019}. \citet{make3010009} show the advantages of the glass box model \textit{Explainable Boosting Machine}, that visualizes feature-wise contributions to the predictions at comparable performance to common models.
To achieve explainability and gain a deeper understanding of what a model has already learned and to avoid spurious correlations~\cite{lapuschkin2019unmasking}, it is best practice to carefully observe the features which impact the model's prediction the most and check whether they are plausible from a domain experts' point of view \cite{bach2015pixel, baehrens2010explain,arras2017relevant}. 
Moreover, case studies have shown that explainability helps to increase trust and users' acceptance \cite{Hois_2019} and could guide humans in ML assisted decisions \cite{schmidt2019quantifying}.
%For example, heat maps highlight the most significant pixels in image classification problems \cite{bach2015pixel, baehrens2010explain} or the most significant words in NLP tasks \cite{arras2017relevant}. 
%For root cause analysis of misclassifications caused by training data issues, the study of \cite{Chakarov_2016} is recommended for further reading. 
%The toolbox by Alber et al.\cite{alber2019innvestigate} provides a unified framework for a wide number of explanation methods.
Unified frameworks to explore model explainabilty are available (e.g. \cite{alber2019innvestigate, nori2019interpretml}).

\textit{Compare results with defined success criteria:} Finally, domain and ML experts have to decide if the model can be deployed. Therefore, it is best practice to document the results of the evaluation phase and compare them to the business and ML success criteria defined in \cref{subsec:success-criteria}. If the success criteria are not met, one might backtrack to earlier phases (modeling or even data preparation) or stop the project. 
Identified limitations of robustness and explainability during evaluation might require an update of the risk assessment (e.g.\ Failure Mode and Effects Analysis (FMEA)) and might also lead to backtracking to the modeling phase or stopping the project.

\subsection{Deployment}\label{Sec:Deployment}
%
%After the model has successfully passed the evaluation state, it is ready to be deployed. 
The deployment phase of a ML model is characterized by its practical use in the designated field of application. 

\textit{Define inference hardware:} Choose the hardware based on the requirements defined in \cref{subsubsec:Assess Situation} or align with an existing hardware. While cloud services offer scalable computation resources, embedded system have hard constraints. ML specific options are e.g. to optimize towards the target hardware \cite{wu2019fb} regarding CPU and GPU availability, to optimize towards the target operation system (demonstrated for Android and iOS by ~\citet{make1010027}) or to optimize the ML workload for a specific platform \cite{workload}. Monitoring and maintenance (see \cref{subsec:Monitoring and Maintenance}) have to be considered in the overall architecture.

%\textcolor{red}{Choose the prediction hardware based on computation, connectivity and business requirments and constrains\textcolor{red}{XXX cite missing}. 
%While cloud services offer a tremendous amount of computation power, a steady, lag free and reliable connection needs to be guaranteed. 
%Complementary, devices at the edge of the cloud have only limited access to large data centers and computations might have to be done locally. }
%Models deployed on embedded systems or even on heterogeneous hardware like Smartphone \cite{wu2019fb} are restricted in size, inference time and updateability and might require approaches to optimize the ML workload (\cite{workload}).
%Offline devices face have to be updated manually or not at all as a consistent connection to a data center can not be ensured.

\textit{Model evaluation under production condition:} 
% words of wisdom not used: Define a strategy how to distribute up-to-date ML models, maintained by the ML deployment team, at regular intervals. 
%As training and test data is gathered to train and evaluate the model, the possible 
The risk persists that the production data does not resemble the training data. %or didn't cover corner cases. 
Previous assumptions on the training data might not hold in production and the hardware that gathered the data might differ. 
Therefore it is best practice to evaluate the performance of the model under incrementally increasing production conditions by iteratively running the tasks in \cref{subsec:Evaluation}. 
On each incremental step, the model has to be calibrated to the deployed hardware and the test environment. 
This allows identifying wrong assumptions on the deployed environment and the causes of model degradation. 
Domain adaptation techniques can be applied \cite{Wang_2018, sugiyama2007covariate} to enhance the generalization ability of the model. This step will also give a first indication whether the business and economic success criteria, which was defined in \cref{subsec:success-criteria}, could be met.
%Face detection algorithms, for example, are trained on still images which allow the ML algorithm to detect key features under controlled conditions. 
%The final test should run the face detection algorithm in real-time on the production hardware, for example, an embedded system, to ensure consistent performance. 

\textit{Assure user acceptance and usability:} Even after passing all evaluation steps, there might be the risk that the user acceptance and the usability of the model is underwhelming. 
The model might be incomprehensible and or does not cover corner cases. 
It is best practice to build a prototype and run an field test with end users \cite{DBLP:journals/corr/abs-1906-10742}.
Examine the acceptance, usage rate and the user experience.
A user guide and disclaimer shall be provided to the end users to explain the system's functionality and limits.

\textit{Minimize the risks of unforeseen errors:} The risks of unforeseen errors and outage times could cause system shutdowns and a temporary suspension of services. 
This could lead to user complaints and the declining of user numbers and could reduce the revenue. 
A fall-back plan, that is activated in case of e.g.\ erroneous model updates or detected bugs, can help to tackle the problem. 
Options are to roll back to a previous version, a pre-defined baseline or a rule-based system. A second option to counteract unforeseen errors is to implement software safety cages that control and limit the outputs of the ML application \cite{heckemann2011safe} or even learn safe regions in the state space \cite{berkenkamp2016safe}.
%Otherwise, it might be necessary to remove the service temporally and reactivate it later on.

\textit{Deployment strategy:} Even though the model is evaluated rigorously during each previous step, there is the risk that errors might be undetected through the process. 
Before rolling out a model, it is best practice to setup an e.g. incremental deployment strategy that includes a pipeline for models and data \cite{modiTFX, derakhshan2019continuous}. When cloud architectures are used, strategies can often be aligned on general deployment strategies for cloud software applications \cite{CloudComputingPatterns2014}.
% \textcolor{red}{deploy it first to a small subset of users and evaluate its behavior in a real-world environment (also called canary deployment) or compare two models in an A/B approach \cite{modiTFX}. }
The impact of such erroneous deployments and the cost of fixing errors should be minimized. 
%If the model successfully passes the canary deployment, it can be deployed to all users.

\subsection{Monitoring and Maintenance} \label{subsec:Monitoring and Maintenance}
With the expansion from knowledge discovery to data-driven applications to infer real-time decisions, ML models are used over a long period and have a life cycle which has to be managed. 
%Maintaining the model assures its quality during its life cycle. 
The risk of not maintaining the model is the degradation of the performance over time which leads to false predictions and could cause errors in subsequent systems. 
The main reason for a model to become impaired over time is rooted in the violation of the assumption that the training data and the input data for inference come from the same distribution. The causes of the violations are:

\begin{itemize}
	\item \textit{Non-stationary data distribution:} Data distributions change over time and result in a stale training set and, thus, the characteristics of the data distribution are represented incorrectly by the training data. Either a shift in the features and/or in the labels are possible. This degrades the performance of the model over time. The frequency of the changes depends on the domain. Data of the stock market are very volatile whereas the visual properties of elephants won't change much over the next years. 
	\item \textit {Degradation of hardware:} The hardware that the model is deployed on and the sensor hardware will age over time. Wear parts in a system will age and friction characteristics of the system might change. Sensors get noisier or fail over time. This will shift the domain of the system and has to be adapted by the model or by retraining it.
	\item \textit{System updates:} Updates on the software or hardware of the system can cause a shift in the environment. For example, the units of a signal got changed during an update. % from kilograms to grams. 
Without notifications, the model would use this scaled input to infer false predictions.
\end{itemize}

After the underlying problem is known, we can formulate the necessary methods to circumvent stale models and assure the quality. We propose two sequential tasks in the maintenance phase to assure or improve the quality of the model. In the \textit{monitor} task, the staleness of the model is evaluated and returns whether the model has to be updated or not. Afterward, the model is updated and evaluated to gauge whether the update was successful.

\textit{Monitor:} Baylor et al.\cite{modiTFX} proposes to monitor all input signals and notify when an update has occurred. Therefore, statistics of the incoming data and the predicted labels can be compared to the statistics of the training data. 
% Updates on the input signals could then be handled automatically or manually. 
Complementary, the schema defined in \cref{subsubsec:Data Quality Verification} can be used to validate the correctness of the incoming data. 
Inputs that do not satisfy the schema can be treated as anomalies and denied by the model~\cite{modiTFX}. Libraries exist to help implementing an automatic data validation system~\cite{schelter2019unit}. 
If the labels of the incoming data are known e.g.\ in forecasting tasks, the performance of the model can be directly monitored and recorded. An equal approach can be applied to the outputs of the model that underlie a certain distribution if environment conditions are stable and can give an estimate on the number of actions performed when interacting with an environment \cite{sculley2015hidden}. The monitoring phase also includes a comparison of the performance with the defined success criteria.
%The results of these data streams could be written in a report and reviewed automatically or manually. 
Based on the monitoring results, it can be decided upon whether the model should be updated e.g.\ if input signals change significantly, the number of anomalies reaches a certain threshold or the performance has reached a lower bound. The decision whether the model has to be updated should consider the costs of updating the model and the costs resulting from erroneous predictions due to stale models. 
%Thresholds are set to notify the system that the model has to be updated and have to be tuned in either case to minimize the update frequency because of the additional overhead but also minimize erroneous predictions due to stale models. 

\textit{Update:} In the updating step, new data is collected to re-train the model under the changed data distribution. 
Consider that new data has to be labeled which could be very expensive. 
Instead of training a completely new model from scratch, it is advised to fine-tune the existing model to new data. 
It might be necessary to perform some of the modeling steps in \cref{subsec:modeling} to cope with the changing data distribution. % e.g.\ by adding additional layers and more weights. 
Every update step has to undergo a new evaluation (\cref{subsec:Evaluation}) before it can be deployed.
%The evaluation tasks in \cref{subsec:Evaluation} are also applied here. 
The performance of the updated model should be compared against the previous versions and could give insights on the time scale of model degradation. 
It should be noted, that ML systems might influence their own behavior during updates due to direct, e.g.\ by influencing its future training data selection, or indirect, e.g.\ via interaction through the world, feedback loops~\cite{sculley2015hidden}. The risk of positive feedback loops causing system instability has to be addressed e.g. by not only monitoring but limiting the actions of the model.

In addition, as part of the deployment strategy, a module is needed that tracks the application usage and performance and handles several deployment strategies like A/B testing \cite{modiTFX, derakhshan2019continuous}. The module can e.g. be set up in form of a microservice \cite{IBMMuthusamy} or a directed graph \cite{ghanta2018interpretability}.
%\cite{IBMMuthusamy} suggest to set up a microservice, that tracks application usage and performance and handels several deployment strategies like canary deployment and A/B testing. 
To reduce the risk of serving erroneous models, an automatic or human controlled fallback to a previous model needs to be implemented. The automation of the update strategy can be boosted up to a continuous training and continuous deployment of the ML application \cite{modiTFX} while covering the defined quality assurance methods.

\section{Discussion}

We have introduced CRISP-ML(Q), a process model for ML applications with quality assurance methodology, that helps organizations to increase efficiency and the success rate in their ML projects. 
It guides ML practitioners through the entire ML development life-cycle, stepping into the phases and tasks of the iterative process including maintenance and monitoring. Whenever tasks specific risks can be identified, we provide quality-oriented methods to mitigate those risks.
%The methods provided are not chosen for their highest degree of novelty, but rather for having passed the test of time to become best practices in automotive industry projects and academia. 
All methods provided are considered best practices in ML projects in industry and academia. % and have the maturity to be implemented in future ML projects.

Our survey is indicative of the existence of specialist literature, but its contributions are not covered in ML textbooks and are not part of the academic curriculum. 
Hence, novices to industry practice often lack a profound state-of-the-art knowledge to mitigate risks and ensure project success. 
Stressing quality assurance methodology is particularly important because many ML practitioners focus solely on improving the predictive performance.

%%%%%%%%%%%%%%%%%%%%%%%%%%%%%%%%%%%%%%%%%%
\section{Conclusions}

An important future step on the basis of our and related work is the standardization of a process model. 
This would contribute to more successful ML projects %in academia and industry 
and thus would have a major impact on the ML community ~\cite{Mariscal_2010}. 

Note that the process and quality measures in this work are not designed for safety-relevant systems. Their study and and the discussion of legal constrains are left to future work.

We encourage industry from automotive and other domains to implement CRISP-ML(Q) in their machine learning applications and contribute their knowledge to establish a CRoss-Industry Standard Process model for the development of machine learning applications with Quality assurance methodology in the future. %Defining the standard is left to future work.

%%%%%%%%%%%%%%%%%%%%%%%%%%%%%%%%%%%%%%%%%%

%%%%%%%%%%%%%%%%%%%%%%%%%%%%%%%%%%%%%%%%%%
\vspace{6pt} 

%%%%%%%%%%%%%%%%%%%%%%%%%%%%%%%%%%%%%%%%%%
%% optional
%\supplementary{The following are available online at \linksupplementary{s1}, Figure S1: title, Table S1: title, Video S1: title.}


%%%%%%%%%%%%%%%%%%%%%%%%%%%%%%%%%%%%%%%%%%
\authorcontributions{Conceptualization, S.St., T.B.B., C.D., A.H., L.W., S.P. and K.-R.M.; methodology, S.St., T.B.B., C.D., A.H., L.W., S.P. and K.-R.M.; writing---original draft preparation, S.St., T.B.B., C.D., A.H., L.W. and S.P.; writing---review and editing, S.St., T.B.B., C.D., A.H., L.W., S.P. and K.-R.M.;  supervision, S.P. and K.-R.M.; project administration, S.St. and T.B.B.; funding acquisition, S.P. and K.-R.M. All authors have read and agreed to the published version of the manuscript.}

\funding{This research was funded by the German Federal Ministry of Education and Research (BMBF) for funding the project \emph{AIAx - Machine Learning-driven Engineering} (Nr. 01IS18048). K.-R. M. acknowledges partial financial support by the BMBF under Grants 01IS14013A-E, 01IS18025A, 01IS18037A, 01GQ1115 and 01GQ0850; Deutsche Forschungsgesellschaft (DFG) under Grant Math+, EXC 2046/1, Project ID 390685689 and by the Technology Promotion (IITP) grant funded by the Korea government (No. 2017-0-00451, No. 2017-0-01779).}

\acknowledgments{Special thanks to the internal Daimler AI community. We would like to thank Miriam H\"agele, Lorenz Linhardt, Simon Letzgus, Danny Panknin and Andreas Ziehe for proofreading the manuscript and the in-depth discussions.}

\conflictsofinterest{The authors declare no conflict of interest. The funders had no role in the writing of the manuscript, or in the decision to publish the~results.} 

\end{paracol}
\reftitle{References}


\externalbibliography{yes}
\bibliography{mybibfile}

\begin{thebibliography}{999}

\bibitem[Lee \em{et~al.}(2015)Lee, Bagheri, and Kao]{lee2015cyber}
Lee, J.; Bagheri, B.; Kao, H.A.
\newblock A cyber-physical systems architecture for industry 4.0-based
  manufacturing systems.
\newblock {\em Manufacturing letters} {\bf 2015}, {\em 3},~18--23.

\bibitem[Brettel \em{et~al.}(2014)Brettel, Friederichsen, Keller, and
  Rosenberg]{brettel2014virtualization}
Brettel, M.; Friederichsen, N.; Keller, M.; Rosenberg, M.
\newblock How virtualization, decentralization and network building change the
  manufacturing landscape: An Industry 4.0 Perspective.
\newblock {\em International journal of mechanical, industrial science and
  engineering} {\bf 2014}, {\em 8},~37--44.

\bibitem[Dikmen and Burns(2016)]{dikmen2016autonomous}
Dikmen, M.; Burns, C.M.
\newblock Autonomous driving in the real world: Experiences with {Tesla}
  autopilot and summon.
\newblock  Proceedings of the 8th international conference on automotive user
  interfaces and interactive vehicular applications. ACM,  2016, pp. 225--228.

\bibitem[Kourou \em{et~al.}(2015)Kourou, Exarchos, Exarchos, Karamouzis, and
  Fotiadis]{kourou2015cancermachinelearning}
Kourou, K.; Exarchos, T.P.; Exarchos, K.P.; Karamouzis, M.V.; Fotiadis, D.I.
\newblock Machine learning applications in cancer prognosis and prediction.
\newblock {\em Computational and structural biotechnology journal} {\bf 2015},
  {\em 13},~8--17.

\bibitem[Esteva \em{et~al.}(2017)Esteva, Kuprel, Novoa, Ko, Swetter, Blau, and
  Thrun]{esteva2017dermatologist}
Esteva, A.; Kuprel, B.; Novoa, R.A.; Ko, J.; Swetter, S.M.; Blau, H.M.; Thrun,
  S.
\newblock Dermatologist-level classification of skin cancer with deep neural
  networks.
\newblock {\em Nature} {\bf 2017}, {\em 542},~115.

\bibitem[Andrews and Hare(2019)]{anha19}
Andrews, W.; Hare, J.
\newblock Survey Analysis: {AI} and {ML} Development Strategies, Motivators and
  Adoption Challenges,  2019.

\bibitem[{Nimdzi Insights}(2019)]{pactera19}
{Nimdzi Insights}.
\newblock Artificial Intelligence: Localization Winners, Losers, Heroes,
  Spectators, and You.
\newblock Technical report, {Pactera EDGE},  2019.

\bibitem[Fischer \em{et~al.}(2021)Fischer, Ehrlinger, Geist, Ramler, Sobiezky,
  Zellinger, Brunner, Kumar, and Moser]{make3010004}
Fischer, L.; Ehrlinger, L.; Geist, V.; Ramler, R.; Sobiezky, F.; Zellinger, W.;
  Brunner, D.; Kumar, M.; Moser, B.
\newblock AI System Engineering—Key Challenges and Lessons Learned.
\newblock {\em Machine Learning and Knowledge Extraction} {\bf 2021}, {\em
  3},~56--83.
\newblock
  doi:{\changeurlcolor{black}\href{https://doi.org/10.3390/make3010004}{\detokenize{10.3390/make3010004}}}.

\bibitem[Hamada \em{et~al.}(2020)Hamada, Ishikawa, Masuda, Matsuya, and
  Ujita]{hamada2020}
Hamada, K.; Ishikawa, F.; Masuda, S.; Matsuya, M.; Ujita, Y.
\newblock Guidelines for quality assurance of machine learning-based artificial
  intelligence.
\newblock  SEKE2020: the 32nd International Conference on Software Engineering
  \& Knowledge Engineering,  2020, pp. 335--341.

\bibitem[Chapman \em{et~al.}(2000)Chapman, Clinton, Kerber, Khabaza, Reinartz,
  Shearer, and Wirth]{crisp-dm}
Chapman, P.; Clinton, J.; Kerber, R.; Khabaza, T.; Reinartz, T.; Shearer, C.;
  Wirth, R.
\newblock {CRISP-DM} 1.0 Step-by-step data mining guide.
\newblock Technical report, The {CRISP-DM} consortium,  2000.

\bibitem[Wirth and Hipp(2000)]{crisp-dm:hipp}
Wirth, R.; Hipp, J.
\newblock {CRISP-DM}: Towards a standard process model for data mining.
\newblock  Proceedings of the Fourth International Conference on the Practical
  Application of Knowledge Discovery and Data Mining,  2000, pp. 29--39.

\bibitem[Shearer(2000)]{crisp-dm:shearer}
Shearer, C.
\newblock The {CRISP-DM} Model: The New Blueprint for Data Mining.
\newblock {\em Journal of Data Warehousing} {\bf 2000}.

\bibitem[Kurgan and Musilek(2006)]{kumu06}
Kurgan, L.; Musilek, P.
\newblock A survey of Knowledge Discovery and Data Mining process models.
\newblock {\em The Knowledge Engineering Review} {\bf 2006}, {\em 21},~1--24.

\bibitem[Mariscal \em{et~al.}(2010)Mariscal, Marbán, and
  Fernández]{Mariscal_2010}
Mariscal, G.; Marbán, O.; Fernández, C.
\newblock A survey of data mining and knowledge discovery process models and
  methodologies.
\newblock {\em Knowledge Eng. Review} {\bf 2010}, {\em 25},~137--166.
\newblock
  doi:{\changeurlcolor{black}\href{https://doi.org/10.1017/S0269888910000032}{\detokenize{10.1017/S0269888910000032}}}.

\bibitem[Kriegel \em{et~al.}(2007)Kriegel, Borgwardt, Kr{\"o}ger, Pryakhin,
  Schubert, and Zimek]{krbokrprsczi07}
Kriegel, H.P.; Borgwardt, K.M.; Kr{\"o}ger, P.; Pryakhin, A.; Schubert, M.;
  Zimek, A.
\newblock Future trends in data mining.
\newblock {\em Data Mining and Knowledge Discovery} {\bf 2007}, {\em
  15},~87--97.

\bibitem[{de Abajo} \em{et~al.}(2004){de Abajo}, Diez, Lobato, and
  Cuesta]{abdilocu04}
{de Abajo}, N.; Diez, A.B.; Lobato, V.; Cuesta, S.R.
\newblock {ANN} Quality Diagnostic Models for Packaging Manufacturing: An
  Industrial Data Mining Case Study.
\newblock  Proceedings of the Tenth ACM SIGKDD International Conference on
  Knowledge Discovery and Data Mining,  2004, pp. 799--804.

\bibitem[Gersten \em{et~al.}(2000)Gersten, Wirth, and Arndt]{gewiar00}
Gersten, W.; Wirth, R.; Arndt, D.
\newblock Predictive modeling in automotive direct marketing: tools,
  experiences and open issues.
\newblock  Proceedings of the Sixth ACM SIGKDD International Conference on
  Knowledge Discovery and Data Mining,  2000, pp. 398--406.

\bibitem[Hipp and Lindner(1999)]{hili99}
Hipp, J.; Lindner, G.
\newblock Analysing Warranty Claims of Automobiles; An Application Description
  following the {CRISP-DM} Data Mining Process.
\newblock  Proceedings of the Fifth International Computer Science Conference,
  1999, pp. 31--40.

\bibitem[IEEE(1997)]{ieee10741997}
IEEE.
\newblock Std 1074-1997, {IEEE} Standard for Developing Software Life Cycle
  Processes.
\newblock Technical report, IEEE,  1997.

\bibitem[Marb\'{a}n \em{et~al.}(2009)Marb\'{a}n, Segovia, Menasalvas, and
  Fern\'{a}ndez-Baiz\'{a}n]{masemefe09}
Marb\'{a}n, O.; Segovia, J.; Menasalvas, E.; Fern\'{a}ndez-Baiz\'{a}n, C.
\newblock Toward data mining engineering: A software engineering approach.
\newblock {\em Information Systems} {\bf 2009}, {\em 34},~87--107.

\bibitem[SAS(2016)]{semma}
SAS.
\newblock {SEMMA} Data Mining Methodology.
\newblock Technical report, SAS Institute,  2016.

\bibitem[Surange(2015)]{surange2015implementation}
Surange, V.G.
\newblock Implementation of {Six Sigma} to reduce cost of quality: a case study
  of automobile sector.
\newblock {\em Journal of Failure Analysis and Prevention} {\bf 2015}, {\em
  15},~282--294.

\bibitem[{Muthusamy} \em{et~al.}(2018){Muthusamy}, {Slominski}, and
  {Ishakian}]{IBMMuthusamy}
{Muthusamy}, V.; {Slominski}, A.; {Ishakian}, V.
\newblock Towards Enterprise-Ready {AI} Deployments Minimizing the Risk of
  Consuming {AI} Models in Business Applications.
\newblock  2018 First International Conference on Artificial Intelligence for
  Industries (AI4I),  2018, pp. 108--109.
\newblock
  doi:{\changeurlcolor{black}\href{https://doi.org/10.1109/AI4I.2018.8665685}{\detokenize{10.1109/AI4I.2018.8665685}}}.

\bibitem[Catley \em{et~al.}(2009)Catley, Smith, McGregor, and
  Tracy]{casmmctr09}
Catley, C.; Smith, K.P.; McGregor, C.; Tracy, M.
\newblock Extending {CRISP-DM} to incorporate temporal data mining of
  multidimensional medical data streams: A neonatal intensive care unit case
  study.
\newblock {\em 22nd IEEE International Symposium on Computer-Based Medical
  Systems} {\bf 2009}, pp. 1--5.

\bibitem[Heath and McGregor(2010)]{hemc10}
Heath, J.; McGregor, C.
\newblock {CRISP-DM0}: A method to extend {CRISP-DM} to support null hypothesis
  driven confirmatory data mining.
\newblock  1st Advances in Health Informatics Conference,  2010, pp. 96--101.

\bibitem[Venter \em{et~al.}(2007)Venter, de~Waal, and Willers]{vewawi07}
Venter, J.; de~Waal, A.; Willers, C.
\newblock Specializing {CRISP-DM} for evidence mining.
\newblock  IFIP International Conference on Digital Forensics. Springer,  2007,
  pp. 303--315.

\bibitem[Niaksu(2015)]{niaksu15}
Niaksu, O.
\newblock {CRISP} Data Mining Methodology Extension for Medical Domain.
\newblock {\em Baltic Journal of Modern Computing} {\bf 2015}, {\em
  3},~92--109.

\bibitem[Amershi \em{et~al.}(2019)Amershi, Begel, Bird, DeLine, Gall, Kamar,
  Nagappan, Nushi, and Zimmermann]{Amershi_2019}
Amershi, S.; Begel, A.; Bird, C.; DeLine, R.; Gall, H.; Kamar, E.; Nagappan,
  N.; Nushi, B.; Zimmermann, T.
\newblock Software Engineering for Machine Learning: A Case Study.
\newblock  International Conference on Software Engineering (ICSE 2019) -
  Software Engineering in Practice track,  2019.

\bibitem[Breck \em{et~al.}(2017)Breck, Cai, Nielsen, Salib, and
  Sculley]{breck2017ml}
Breck, E.; Cai, S.; Nielsen, E.; Salib, M.; Sculley, D.
\newblock The {ML} test score: A rubric for {ML} production readiness and
  technical debt reduction.
\newblock  2017 IEEE International Conference on Big Data (Big Data). IEEE,
  2017, pp. 1123--1132.

\bibitem[Sculley \em{et~al.}(2015)Sculley, Holt, Golovin, Davydov, Phillips,
  Ebner, Chaudhary, Young, Crespo, and Dennison]{sculley2015hidden}
Sculley, D.; Holt, G.; Golovin, D.; Davydov, E.; Phillips, T.; Ebner, D.;
  Chaudhary, V.; Young, M.; Crespo, J.F.; Dennison, D.
\newblock Hidden technical debt in machine learning systems.
\newblock  Advances in neural information processing systems,  2015, pp.
  2503--2511.

\bibitem[Falcini \em{et~al.}(2017)Falcini, Lami, and
  Mitidieri~Costanza]{wmodel}
Falcini, F.; Lami, G.; Mitidieri~Costanza, A.
\newblock Deep Learning in Automotive Software.
\newblock {\em IEEE Software} {\bf 2017}, {\em 34},~56--63.
\newblock
  doi:{\changeurlcolor{black}\href{https://doi.org/10.1109/MS.2017.79}{\detokenize{10.1109/MS.2017.79}}}.

\bibitem[Kuwajima \em{et~al.}(2018)Kuwajima, Yasuoka, and Nakae]{denso}
Kuwajima, H.; Yasuoka, H.; Nakae, T.
\newblock Open Problems in Engineering and Quality Assurance of Safety Critical
  Machine Learning Systems.
\newblock {\em CoRR} {\bf 2018}, {\em abs/1812.03057},
  \href{http://xxx.lanl.gov/abs/1812.03057}{{\normalfont [1812.03057]}}.

\bibitem[Watanabe \em{et~al.}(2019)Watanabe, Washizaki, Sakamoto, Saito, Honda,
  Tsuda, Fukazawa, and Yoshioka]{watanabe2019preliminary}
Watanabe, Y.; Washizaki, H.; Sakamoto, K.; Saito, D.; Honda, K.; Tsuda, N.;
  Fukazawa, Y.; Yoshioka, N.
\newblock Preliminary Systematic Literature Review of Machine Learning System
  Development Process,  2019,
  \href{http://xxx.lanl.gov/abs/1910.05528}{{\normalfont
  [arXiv:cs.LG/1910.05528]}}.

\bibitem[Rudin and Carlson(2019)]{Rudin_2019_secrets}
Rudin, C.; Carlson, D.
\newblock The Secrets of Machine Learning: Ten Things You Wish You Had Known
  Earlier to be More Effective at Data Analysis,  2019,
  \href{http://xxx.lanl.gov/abs/1906.01998}{{\normalfont
  [arXiv:cs.LG/1906.01998]}}.

\bibitem[Pudaruth(2014)]{pudaruth2014predicting}
Pudaruth, S.
\newblock Predicting the price of used cars using machine learning techniques.
\newblock {\em Int. J. Inf. Comput. Technol} {\bf 2014}, {\em 4},~753--764.

\bibitem[Reed \em{et~al.}(2016)Reed, Kennedy, and
  Silva]{reed2016responsibility}
Reed, C.; Kennedy, E.; Silva, S.
\newblock Responsibility, Autonomy and Accountability: legal liability for
  machine learning.
\newblock {\em Queen Mary School of Law Legal Studies Research Paper} {\bf
  2016}.

\bibitem[Bibal \em{et~al.}(2020)Bibal, Lognoul, de~Streel, and
  Fr{\'e}nay]{bibal2020legal}
Bibal, A.; Lognoul, M.; de~Streel, A.; Fr{\'e}nay, B.
\newblock Legal requirements on explainability in machine learning.
\newblock {\em Artificial Intelligence and Law} {\bf 2020}, pp. 1--21.

\bibitem[Binns(2018)]{Binns_2018}
Binns, R.
\newblock Fairness in Machine Learning: Lessons from Political Philosophy.
\newblock  Proceedings of the 1st Conference on Fairness, Accountability and
  Transparency; Friedler, S.A.; Wilson, C., Eds.; PMLR: New York, NY, USA,
  2018; Vol.~81, {\em Proceedings of Machine Learning Research}, pp. 149--159.

\bibitem[Corbett-Davies and Goel(2018)]{Corbett-Davies_2018}
Corbett-Davies, S.; Goel, S.
\newblock The Measure and Mismeasure of Fairness: A Critical Review of Fair
  Machine Learning,  2018,
  \href{http://xxx.lanl.gov/abs/1808.00023}{{\normalfont
  [arXiv:cs.CY/1808.00023]}}.

\bibitem[{Barredo Arrieta} \em{et~al.}(2019){Barredo Arrieta},
  {D{\'\i}az-Rodr{\'\i}guez}, {Del Ser}, {Bennetot}, {Tabik}, {Barbado},
  {Garc{\'\i}a}, {Gil-L{\'o}pez}, {Molina}, {Benjamins}, {Chatila}, and
  {Herrera}]{Arrieta_2019}
{Barredo Arrieta}, A.; {D{\'\i}az-Rodr{\'\i}guez}, N.; {Del Ser}, J.;
  {Bennetot}, A.; {Tabik}, S.; {Barbado}, A.; {Garc{\'\i}a}, S.;
  {Gil-L{\'o}pez}, S.; {Molina}, D.; {Benjamins}, R.; {Chatila}, R.; {Herrera},
  F.
\newblock {Explainable Artificial Intelligence (XAI): Concepts, Taxonomies,
  Opportunities and Challenges toward Responsible AI} {\bf 2019}.
\newblock  \href{http://xxx.lanl.gov/abs/1910.10045}{{\normalfont
  [arXiv:cs.AI/1910.10045]}}.

\bibitem[McQueen \em{et~al.}(2016)McQueen, Meil{\u{a}}, VanderPlas, and
  Zhang]{mcqueen2016megaman}
McQueen, J.; Meil{\u{a}}, M.; VanderPlas, J.; Zhang, Z.
\newblock Megaman: scalable manifold learning in python.
\newblock {\em The Journal of Machine Learning Research} {\bf 2016}, {\em
  17},~5176--5180.

\bibitem[Polyzotis \em{et~al.}(2017)Polyzotis, Roy, Whang, and
  Zinkevich]{poly17}
Polyzotis, N.; Roy, S.; Whang, S.E.; Zinkevich, M.
\newblock Data management challenges in production machine learning.
\newblock  Proceedings of the 2017 ACM International Conference on Management
  of Data. ACM,  2017, pp. 1723--1726.

\bibitem[Schelter \em{et~al.}(2019)Schelter, Biessmann, Lange, Rukat, Schmidt,
  Seufert, Brunelle, and Taptunov]{schelter2019unit}
Schelter, S.; Biessmann, F.; Lange, D.; Rukat, T.; Schmidt, P.; Seufert, S.;
  Brunelle, P.; Taptunov, A.
\newblock Unit Testing Data with {Deequ}.
\newblock  Proceedings of the 2019 International Conference on Management of
  Data. ACM,  2019, pp. 1993--1996.

\bibitem[Keogh and Mueen(2017)]{Keogh_2017}
Keogh, E.; Mueen, A., Curse of Dimensionality.
\newblock In {\em Encyclopedia of Machine Learning and Data Mining}; Sammut,
  C.; Webb, G.I., Eds.; Springer US: Boston, MA,  2017; pp. 314--315.
\newblock
  doi:{\changeurlcolor{black}\href{https://doi.org/10.1007/978-1-4899-7687-1_192}{\detokenize{10.1007/978-1-4899-7687-1_192}}}.

\bibitem[Bishop(2007)]{Bishop_2007}
Bishop, C.M.
\newblock {\em Pattern recognition and machine learning, 5th Edition};
  Information science and statistics, Springer,  2007.

\bibitem[Guyon and Elisseeff(2003)]{Guyon:2003:IVF:944919.944968}
Guyon, I.; Elisseeff, A.
\newblock An Introduction to Variable and Feature Selection.
\newblock {\em J. Mach. Learn. Res.} {\bf 2003}, {\em 3},~1157--1182.

\bibitem[Braun \em{et~al.}(2008)Braun, Buhmann, and
  M{\"u}ller]{braun2008relevant}
Braun, M.L.; Buhmann, J.M.; M{\"u}ller, K.R.
\newblock On relevant dimensions in kernel feature spaces.
\newblock {\em Journal of Machine Learning Research} {\bf 2008}, {\em
  9},~1875--1908.

\bibitem[Hira and Gillies(2015)]{hira2015review}
Hira, Z.M.; Gillies, D.F.
\newblock A review of feature selection and feature extraction methods applied
  on microarray data.
\newblock {\em Advances in bioinformatics} {\bf 2015}, {\em 2015}.

\bibitem[Saeys \em{et~al.}(2007)Saeys, Inza, and
  Larra\~{n}aga]{Saeys:2007:RFS:1349154.1349169}
Saeys, Y.; Inza, I.n.; Larra\~{n}aga, P.
\newblock A Review of Feature Selection Techniques in Bioinformatics.
\newblock {\em Bioinformatics} {\bf 2007}, {\em 23},~2507--2517.
\newblock
  doi:{\changeurlcolor{black}\href{https://doi.org/10.1093/bioinformatics/btm344}{\detokenize{10.1093/bioinformatics/btm344}}}.

\bibitem[Chandrashekar and Sahin(2014)]{Chandrashekar:2014:SFS:2577586.2577699}
Chandrashekar, G.; Sahin, F.
\newblock A Survey on Feature Selection Methods.
\newblock {\em Comput. Electr. Eng.} {\bf 2014}, {\em 40},~16--28.
\newblock
  doi:{\changeurlcolor{black}\href{https://doi.org/10.1016/j.compeleceng.2013.11.024}{\detokenize{10.1016/j.compeleceng.2013.11.024}}}.

\bibitem[Guyon \em{et~al.}(2006)Guyon, Gunn, Nikravesh, and
  Zadeh]{Guyon:2006:FEF:1208773}
Guyon, I.; Gunn, S.; Nikravesh, M.; Zadeh, L.A.
\newblock {\em Feature Extraction: Foundations and Applications (Studies in
  Fuzziness and Soft Computing)}; Springer-Verlag: Berlin, Heidelberg,  2006.

\bibitem[Ambroise and McLachlan(2002)]{ambroise2002selection}
Ambroise, C.; McLachlan, G.J.
\newblock Selection bias in gene extraction on the basis of microarray
  gene-expression data.
\newblock {\em Proceedings of the national academy of sciences} {\bf 2002},
  {\em 99},~6562--6566.

\bibitem[Lapuschkin \em{et~al.}(2016)Lapuschkin, Binder, Montavon, M{\"u}ller,
  and Samek]{Lapuschkin_FisherVectorLRP}
Lapuschkin, S.; Binder, A.; Montavon, G.; M{\"u}ller, K.R.; Samek, W.
\newblock Analyzing classifiers: Fisher vectors and deep neural networks.
\newblock  Proceedings of the IEEE Conference on Computer Vision and Pattern
  Recognition,  2016, pp. 2912--2920.

\bibitem[Lapuschkin \em{et~al.}(2019)Lapuschkin, W{\"a}ldchen, Binder,
  Montavon, Samek, and M{\"u}ller]{lapuschkin2019unmasking}
Lapuschkin, S.; W{\"a}ldchen, S.; Binder, A.; Montavon, G.; Samek, W.;
  M{\"u}ller, K.R.
\newblock Unmasking {Clever Hans} predictors and assessing what machines really
  learn.
\newblock {\em Nature communications} {\bf 2019}, {\em 10},~1096.

\bibitem[Samek \em{et~al.}(2019)Samek, Montavon, Vedaldi, Hansen, and
  M{\"u}ller]{samek2019explainable}
Samek, W.; Montavon, G.; Vedaldi, A.; Hansen, L.K.; M{\"u}ller, K.R.
\newblock {\em Explainable AI: interpreting, explaining and visualizing deep
  learning}; Vol. 11700, Springer Nature,  2019.

\bibitem[Lawrence \em{et~al.}(1998)Lawrence, Burns, Back, Tsoi, and
  Giles]{lawrence1998neural}
Lawrence, S.; Burns, I.; Back, A.; Tsoi, A.C.; Giles, C.L.
\newblock Neural network classification and prior class probabilities. In {\em
  Neural networks: tricks of the trade}; Springer,  1998; pp. 299--313.

\bibitem[Chawla \em{et~al.}(2002)Chawla, Bowyer, Hall, and
  Kegelmeyer]{Chawla_2002}
Chawla, N.V.; Bowyer, K.W.; Hall, L.O.; Kegelmeyer, W.P.
\newblock SMOTE: Synthetic Minority Over-sampling Technique.
\newblock {\em Journal of Artificial Intelligence Research} {\bf 2002}, {\em
  16},~321--357.

\bibitem[Batista \em{et~al.}(2004)Batista, Prati, and Monard]{Batista_2004}
Batista, G.E.A.P.A.; Prati, R.C.; Monard, M.C.
\newblock A Study of the Behavior of Several Methods for Balancing Machine
  Learning Training Data.
\newblock {\em SIGKDD Explor. Newsl.} {\bf 2004}, {\em 6},~20--29.
\newblock
  doi:{\changeurlcolor{black}\href{https://doi.org/10.1145/1007730.1007735}{\detokenize{10.1145/1007730.1007735}}}.

\bibitem[Lema\^{\i}tre \em{et~al.}(2017)Lema\^{\i}tre, Nogueira, and
  Aridas]{Lemaitre:2017:IPT:3122009.3122026}
Lema\^{\i}tre, G.; Nogueira, F.; Aridas, C.K.
\newblock Imbalanced-learn: A {Python} Toolbox to Tackle the Curse of
  Imbalanced Datasets in Machine Learning.
\newblock {\em J. Mach. Learn. Res.} {\bf 2017}, {\em 18},~559--563.

\bibitem[Walker(2002)]{walker2002primer}
Walker, J.S.
\newblock {\em A primer on wavelets and their scientific applications}; CRC
  press,  2002.

\bibitem[Lyons(2004)]{Lyons:2004:UDS:993484}
Lyons, R.G.
\newblock {\em Understanding Digital Signal Processing (2Nd Edition)}; Prentice
  Hall PTR: Upper Saddle River, NJ, USA,  2004.

\bibitem[Che \em{et~al.}(2018)Che, Purushotham, Cho, Sontag, and
  Liu]{che2018recurrent}
Che, Z.; Purushotham, S.; Cho, K.; Sontag, D.; Liu, Y.
\newblock Recurrent neural networks for multivariate time series with missing
  values.
\newblock {\em Scientific reports} {\bf 2018}, {\em 8},~6085.

\bibitem[Biessmann \em{et~al.}(2018)Biessmann, Salinas, Schelter, Schmidt, and
  Lange]{biessmann2018deep}
Biessmann, F.; Salinas, D.; Schelter, S.; Schmidt, P.; Lange, D.
\newblock Deep Learning for Missing Value Imputationin Tables with
  Non-Numerical Data.
\newblock  Proceedings of the 27th ACM International Conference on Information
  and Knowledge Management. ACM,  2018, pp. 2017--2025.

\bibitem[Koren \em{et~al.}(2009)Koren, Bell, and
  Volinsky]{Koren:2009:MFT:1608565.1608614}
Koren, Y.; Bell, R.; Volinsky, C.
\newblock Matrix Factorization Techniques for Recommender Systems.
\newblock {\em Computer} {\bf 2009}, {\em 42},~30--37.
\newblock
  doi:{\changeurlcolor{black}\href{https://doi.org/10.1109/MC.2009.263}{\detokenize{10.1109/MC.2009.263}}}.

\bibitem[Murray \em{et~al.}(2018)Murray et~al.]{murray2018multiple}
Murray, J.S.; others.
\newblock Multiple imputation: a review of practical and theoretical findings.
\newblock {\em Statistical Science} {\bf 2018}, {\em 33},~142--159.

\bibitem[White \em{et~al.}(2011)White, Royston, and Wood]{white2011multiple}
White, I.R.; Royston, P.; Wood, A.M.
\newblock Multiple imputation using chained equations: issues and guidance for
  practice.
\newblock {\em Statistics in medicine} {\bf 2011}, {\em 30},~377--399.

\bibitem[Azur \em{et~al.}(2011)Azur, Stuart, Frangakis, and
  Leaf]{azur2011multiple}
Azur, M.J.; Stuart, E.A.; Frangakis, C.; Leaf, P.J.
\newblock Multiple imputation by chained equations: what is it and how does it
  work?
\newblock {\em International journal of methods in psychiatric research} {\bf
  2011}, {\em 20},~40--49.

\bibitem[Bertsimas \em{et~al.}(2018)Bertsimas, Pawlowski, and
  Zhuo]{JMLR:v18:17-073}
Bertsimas, D.; Pawlowski, C.; Zhuo, Y.D.
\newblock From Predictive Methods to Missing Data Imputation: An Optimization
  Approach.
\newblock {\em Journal of Machine Learning Research} {\bf 2018}, {\em
  18},~1--39.

\bibitem[Coates and Ng(2012)]{coates2012learning}
Coates, A.; Ng, A.Y.
\newblock Learning feature representations with k-means. In {\em Neural
  networks: Tricks of the trade}; Springer,  2012; pp. 561--580.

\bibitem[Sch{\"o}lkopf \em{et~al.}(1997)Sch{\"o}lkopf, Smola, and
  M{\"u}ller]{scholkopf1997kernel}
Sch{\"o}lkopf, B.; Smola, A.; M{\"u}ller, K.R.
\newblock Kernel principal component analysis.
\newblock  International conference on artificial neural networks. Springer,
  1997, pp. 583--588.

\bibitem[Rumelhart \em{et~al.}(1985)Rumelhart, Hinton, and
  Williams]{rumelhart1985learning}
Rumelhart, D.E.; Hinton, G.E.; Williams, R.J.
\newblock Learning internal representations by error propagation.
\newblock Technical report, California Univ San Diego La Jolla Inst for
  Cognitive Science,  1985.

\bibitem[Goodfellow \em{et~al.}(2016)Goodfellow, Bengio, and
  Courville]{Goodfellow_2016}
Goodfellow, I.; Bengio, Y.; Courville, A.
\newblock {\em Deep Learning}; MIT Press,  2016.

\bibitem[Wong \em{et~al.}(2016)Wong, Gatt, Stamatescu, and
  McDonnell]{wong2016understanding}
Wong, S.C.; Gatt, A.; Stamatescu, V.; McDonnell, M.D.
\newblock Understanding data augmentation for classification: when to warp?
\newblock  2016 international conference on digital image computing: techniques
  and applications (DICTA). IEEE,  2016, pp. 1--6.

\bibitem[LeCun \em{et~al.}(2012)LeCun, Bottou, Orr, and
  M{\"u}ller]{lecun2012efficient}
LeCun, Y.A.; Bottou, L.; Orr, G.B.; M{\"u}ller, K.R.
\newblock Efficient backprop. In {\em Neural networks: Tricks of the trade};
  Springer,  2012; pp. 9--48.

\bibitem[Ioffe and Szegedy(2015)]{DBLP:journals/corr/IoffeS15}
Ioffe, S.; Szegedy, C.
\newblock Batch Normalization: Accelerating Deep Network Training by Reducing
  Internal Covariate Shift.
\newblock {\em CoRR} {\bf 2015}, {\em abs/1502.03167},
  \href{http://xxx.lanl.gov/abs/1502.03167}{{\normalfont [1502.03167]}}.

\bibitem[Baylor \em{et~al.}(2017)Baylor, Breck, Cheng, Fiedel, Foo, Haque,
  Haykal, Ispir, Jain, Koc, et~al.]{modiTFX}
Baylor, D.; Breck, E.; Cheng, H.T.; Fiedel, N.; Foo, C.Y.; Haque, Z.; Haykal,
  S.; Ispir, M.; Jain, V.; Koc, L.; others.
\newblock {TFX}: A {TensorFlow}-based production-scale machine learning
  platform.
\newblock  Proceedings of the 23rd ACM SIGKDD International Conference on
  Knowledge Discovery and Data Mining,  2017, pp. 1387--1395.

\bibitem[Rudin(2019)]{Rudin_2019}
Rudin, C.
\newblock Stop explaining black box machine learning models for high stakes
  decisions and use interpretable models instead.
\newblock {\em Nature Machine Intelligence} {\bf 2019}, {\em 1}.

\bibitem[Schmidt and Biessmann(2019)]{schmidt2019quantifying}
Schmidt, P.; Biessmann, F.
\newblock Quantifying Interpretability and Trust in Machine Learning Systems.
\newblock {\em arXiv preprint arXiv:1901.08558} {\bf 2019}.

\bibitem[Sch{\"o}lkopf and Smola(2002)]{scholkopf2002learning}
Sch{\"o}lkopf, B.; Smola, A.J.
\newblock {\em Learning with kernels: support vector machines, regularization,
  optimization, and beyond}; MIT press,  2002.

\bibitem[Wolpert(1996)]{Wolpert:1996:LPD:1362127.1362128}
Wolpert, D.H.
\newblock The Lack of a Priori Distinctions Between Learning Algorithms.
\newblock {\em Neural Comput.} {\bf 1996}, {\em 8},~1341--1390.
\newblock
  doi:{\changeurlcolor{black}\href{https://doi.org/10.1162/neco.1996.8.7.1341}{\detokenize{10.1162/neco.1996.8.7.1341}}}.

\bibitem[M{\"u}ller \em{et~al.}(2001)M{\"u}ller, Mika, R{\"a}tsch, Tsuda, and
  Sch{\"o}lkopf]{muller2001introduction}
M{\"u}ller, K.R.; Mika, S.; R{\"a}tsch, G.; Tsuda, K.; Sch{\"o}lkopf, B.
\newblock An introduction to kernel-based learning algorithms.
\newblock {\em IEEE transactions on neural networks} {\bf 2001}, {\em
  12},~181--201.

\bibitem[Zhang \em{et~al.}(2019)Zhang, Harman, Ma, and
  Liu]{DBLP:journals/corr/abs-1906-10742}
Zhang, J.M.; Harman, M.; Ma, L.; Liu, Y.
\newblock Machine Learning Testing: Survey, Landscapes and Horizons.
\newblock {\em CoRR} {\bf 2019}, {\em abs/1906.10742},
  \href{http://xxx.lanl.gov/abs/1906.10742}{{\normalfont [1906.10742]}}.

\bibitem[Hutter \em{et~al.}(2019)Hutter, Kotthoff, and Vanschoren]{Hutter_2019}
Hutter, F.; Kotthoff, L.; Vanschoren, J., Eds.
\newblock {\em Automated Machine Learning - Methods, Systems, Challenges}; The
  Springer Series on Challenges in Machine Learning, Springer,  2019.
\newblock
  doi:{\changeurlcolor{black}\href{https://doi.org/10.1007/978-3-030-05318-5}{\detokenize{10.1007/978-3-030-05318-5}}}.

\bibitem[Feurer \em{et~al.}(2015)Feurer, Klein, Eggensperger, Springenberg,
  Blum, and Hutter]{Feurer_2015}
Feurer, M.; Klein, A.; Eggensperger, K.; Springenberg, J.; Blum, M.; Hutter, F.
\newblock Efficient and Robust Automated Machine Learning. In {\em Advances in
  Neural Information Processing Systems 28}; Cortes, C.; Lawrence, N.D.; Lee,
  D.D.; Sugiyama, M.; Garnett, R., Eds.; Curran Associates, Inc.,  2015; pp.
  2962--2970.

\bibitem[Zoph and Le(2016)]{zoph2016neural}
Zoph, B.; Le, Q.V.
\newblock Neural architecture search with reinforcement learning.
\newblock {\em arXiv preprint arXiv:1611.01578} {\bf 2016}.

\bibitem[Erhan \em{et~al.}(2010)Erhan, Bengio, Courville, Manzagol, Vincent,
  and Bengio]{Erhan:2010:WUP:1756006.1756025}
Erhan, D.; Bengio, Y.; Courville, A.; Manzagol, P.A.; Vincent, P.; Bengio, S.
\newblock Why Does Unsupervised Pre-training Help Deep Learning?
\newblock {\em J. Mach. Learn. Res.} {\bf 2010}, {\em 11},~625--660.

\bibitem[Dreher \em{et~al.}(2020)Dreher, Schmidt, Welch, Ourza, Z\"undorf,
  Maucher, Peters, Dreizler, Böhm, and Hanuschkin]{Dreher_2020}
Dreher, D.; Schmidt, M.; Welch, C.; Ourza, S.; Z\"undorf, S.; Maucher, J.;
  Peters, S.; Dreizler, A.; Böhm, B.; Hanuschkin, A.
\newblock Deep Feature Learning of In-Cylinder Flow Fields to Analyze {CCVs} in
  an {SI}-Engine.
\newblock {\em International Journal of Engine Research} {\bf 2020}.
\newblock
  doi:{\changeurlcolor{black}\href{https://doi.org/10.1177/1468087420974148}{\detokenize{10.1177/1468087420974148}}}.

\bibitem[Kingma \em{et~al.}(2014)Kingma, Mohamed, Jimenez~Rezende, and
  Welling]{NIPS2014_5352}
Kingma, D.P.; Mohamed, S.; Jimenez~Rezende, D.; Welling, M.
\newblock Semi-supervised Learning with Deep Generative Models. In {\em
  Advances in Neural Information Processing Systems 27}; Ghahramani, Z.;
  Welling, M.; Cortes, C.; Lawrence, N.D.; Weinberger, K.Q., Eds.; Curran
  Associates, Inc.,  2014; pp. 3581--3589.

\bibitem[Chapelle \em{et~al.}(2010)Chapelle, Schlkopf, and
  Zien]{Chapelle:2010:SL:1841234}
Chapelle, O.; Schlkopf, B.; Zien, A.
\newblock {\em Semi-Supervised Learning}, 1st ed.; The MIT Press,  2010.

\bibitem[Yosinski \em{et~al.}(2014)Yosinski, Clune, Bengio, and
  Lipson]{NIPS2014_5347}
Yosinski, J.; Clune, J.; Bengio, Y.; Lipson, H.
\newblock How transferable are features in deep neural networks? In {\em
  Advances in Neural Information Processing Systems 27}; Ghahramani, Z.;
  Welling, M.; Cortes, C.; Lawrence, N.D.; Weinberger, K.Q., Eds.; Curran
  Associates, Inc.,  2014; pp. 3320--3328.

\bibitem[Williams and Seeger(2001)]{NIPS2000_1866}
Williams, C.K.I.; Seeger, M.
\newblock Using the {N}ystr\"{o}m Method to Speed Up Kernel Machines. In {\em
  Advances in Neural Information Processing Systems 13}; Leen, T.K.;
  Dietterich, T.G.; Tresp, V., Eds.; MIT Press,  2001; pp. 682--688.

\bibitem[Drineas and Mahoney(2005)]{Drineas:2005:NMA:1046920.1194916}
Drineas, P.; Mahoney, M.W.
\newblock On the {N}ystr\"{o}m Method for Approximating a {G}ram Matrix for
  Improved Kernel-Based Learning.
\newblock {\em J. Mach. Learn. Res.} {\bf 2005}, {\em 6},~2153--2175.

\bibitem[Cheng \em{et~al.}(2017)Cheng, Wang, Zhou, and
  Zhang]{DBLP:journals/corr/abs-1710-09282}
Cheng, Y.; Wang, D.; Zhou, P.; Zhang, T.
\newblock A Survey of Model Compression and Acceleration for Deep Neural
  Networks.
\newblock {\em CoRR} {\bf 2017}, {\em abs/1710.09282},
  \href{http://xxx.lanl.gov/abs/1710.09282}{{\normalfont [1710.09282]}}.

\bibitem[Frankle and Carbin(2018)]{frankle2018the}
Frankle, J.; Carbin, M.
\newblock The lottery ticket hypothesis: Finding sparse, trainable neural
  networks.
\newblock {\em arXiv preprint arXiv:1803.03635} {\bf 2018}.

\bibitem[Wiedemann \em{et~al.}(2019)Wiedemann, Kirchhoffer, Matlage, Haase,
  Marb{\'{a}}n, Marinc, Neumann, Nguyen, Osman, Marpe, Schwarz, Wiegand, and
  Samek]{DBLP:journals/corr/abs-1907-11900}
Wiedemann, S.; Kirchhoffer, H.; Matlage, S.; Haase, P.; Marb{\'{a}}n, A.;
  Marinc, T.; Neumann, D.; Nguyen, T.; Osman, A.; Marpe, D.; Schwarz, H.;
  Wiegand, T.; Samek, W.
\newblock DeepCABAC: {A} Universal Compression Algorithm for Deep Neural
  Networks.
\newblock {\em CoRR} {\bf 2019}, {\em abs/1907.11900},
  \href{http://xxx.lanl.gov/abs/1907.11900}{{\normalfont [1907.11900]}}.

\bibitem[Rokach(2010)]{rokach2010ensemble}
Rokach, L.
\newblock Ensemble-based classifiers.
\newblock {\em Artificial Intelligence Review} {\bf 2010}, {\em 33},~1--39.

\bibitem[Zhou \em{et~al.}(2002)Zhou, Wu, and Tang]{zhou2002ensembling}
Zhou, Z.H.; Wu, J.; Tang, W.
\newblock Ensembling neural networks: many could be better than all.
\newblock {\em Artificial intelligence} {\bf 2002}, {\em 137},~239--263.

\bibitem[Opitz and Maclin(1999)]{opitz1999popular}
Opitz, D.; Maclin, R.
\newblock Popular ensemble methods: An empirical study.
\newblock {\em Journal of artificial intelligence research} {\bf 1999}, {\em
  11},~169--198.

\bibitem[Lakshminarayanan \em{et~al.}(2017)Lakshminarayanan, Pritzel, and
  Blundell]{lakshminarayanan2017simple}
Lakshminarayanan, B.; Pritzel, A.; Blundell, C.
\newblock Simple and scalable predictive uncertainty estimation using deep
  ensembles.
\newblock  Advances in Neural Information Processing Systems,  2017, pp.
  6402--6413.

\bibitem[Gal and Ghahramani(2016)]{gal2016dropout}
Gal, Y.; Ghahramani, Z.
\newblock Dropout as a {Bayesian} approximation: Representing model uncertainty
  in deep learning.
\newblock  international conference on machine learning,  2016, pp. 1050--1059.

\bibitem[Pineau(2019)]{Pineau_2019}
Pineau, J.
\newblock The Machine Learning Reproducibility Checklist,  2019.
\newblock Accessed: 2019-06-11.

\bibitem[Tatman \em{et~al.}(2018)Tatman, VanderPlas, and
  Dane]{tatman2018practical}
Tatman, R.; VanderPlas, J.; Dane, S.
\newblock A Practical Taxonomy of Reproducibility for Machine Learning Research
  {\bf 2018}.

\bibitem[Henderson \em{et~al.}(2018)Henderson, Islam, Bachman, Pineau, Precup,
  and Meger]{henderson2018deep}
Henderson, P.; Islam, R.; Bachman, P.; Pineau, J.; Precup, D.; Meger, D.
\newblock Deep reinforcement learning that matters.
\newblock  Thirty-Second AAAI Conference on Artificial Intelligence,  2018.

\bibitem[Sculley \em{et~al.}(2018)Sculley, Snoek, Wiltschko, and
  Rahimi]{sculley2018winner's}
Sculley, D.; Snoek, J.; Wiltschko, A.; Rahimi, A.
\newblock Winner's Curse? On Pace, Progress, and Empirical Rigor,  2018.

\bibitem[Bouthillier \em{et~al.}(2019)Bouthillier, Laurent, and
  Vincent]{pmlr-v97-bouthillier19a}
Bouthillier, X.; Laurent, C.; Vincent, P.
\newblock Unreproducible Research is Reproducible.
\newblock  Proceedings of the 36th International Conference on Machine
  Learning; Chaudhuri, K.; Salakhutdinov, R., Eds.; PMLR: Long Beach,
  California, USA,  2019; Vol.~97, {\em Proceedings of Machine Learning
  Research}, pp. 725--734.

\bibitem[Vartak \em{et~al.}(2016)Vartak, Subramanyam, Lee, Viswanathan, Husnoo,
  Madden, and Zaharia]{Vartak}
Vartak, M.; Subramanyam, H.; Lee, W.E.; Viswanathan, S.; Husnoo, S.; Madden,
  S.; Zaharia, M.
\newblock ModelDB: A System for Machine Learning Model Management.
\newblock  Proceedings of the Workshop on Human-In-the-Loop Data Analytics;
  ACM: New York, NY, USA,  2016; HILDA '16, pp. 14:1--14:3.
\newblock
  doi:{\changeurlcolor{black}\href{https://doi.org/10.1145/2939502.2939516}{\detokenize{10.1145/2939502.2939516}}}.

\bibitem[Zhou and Sun(2019)]{Zhou:2019:MTD:3314328.3241979}
Zhou, Z.Q.; Sun, L.
\newblock Metamorphic Testing of Driverless Cars.
\newblock {\em Commun. ACM} {\bf 2019}, {\em 62},~61--67.
\newblock
  doi:{\changeurlcolor{black}\href{https://doi.org/10.1145/3241979}{\detokenize{10.1145/3241979}}}.

\bibitem[Tian \em{et~al.}(2018)Tian, Pei, Jana, and
  Ray]{Tian:2018:DAT:3180155.3180220}
Tian, Y.; Pei, K.; Jana, S.; Ray, B.
\newblock {DeepTest}: Automated Testing of Deep-neural-network-driven
  Autonomous Cars.
\newblock  Proceedings of the 40th International Conference on Software
  Engineering; ACM: New York, NY, USA,  2018; ICSE '18, pp. 303--314.
\newblock
  doi:{\changeurlcolor{black}\href{https://doi.org/10.1145/3180155.3180220}{\detokenize{10.1145/3180155.3180220}}}.

\bibitem[Pei \em{et~al.}(2017)Pei, Cao, Yang, and
  Jana]{Pei:2017:DAW:3132747.3132785}
Pei, K.; Cao, Y.; Yang, J.; Jana, S.
\newblock {DeepXplore}: Automated Whitebox Testing of Deep Learning Systems.
\newblock  Proceedings of the 26th Symposium on Operating Systems Principles;
  ACM: New York, NY, USA,  2017; SOSP '17, pp. 1--18.
\newblock
  doi:{\changeurlcolor{black}\href{https://doi.org/10.1145/3132747.3132785}{\detokenize{10.1145/3132747.3132785}}}.

\bibitem[Chan-Hon-Tong(2019)]{make1010011}
Chan-Hon-Tong, A.
\newblock An Algorithm for Generating Invisible Data Poisoning Using
  Adversarial Noise That Breaks Image Classification Deep Learning.
\newblock {\em Machine Learning and Knowledge Extraction} {\bf 2019}, {\em
  1},~192--204.
\newblock
  doi:{\changeurlcolor{black}\href{https://doi.org/10.3390/make1010011}{\detokenize{10.3390/make1010011}}}.

\bibitem[Chakarov \em{et~al.}(2016)Chakarov, Nori, Rajamani, Sen, and
  Vijaykeerthy]{Chakarov_2016}
Chakarov, A.; Nori, A.V.; Rajamani, S.K.; Sen, S.; Vijaykeerthy, D.
\newblock Debugging Machine Learning Tasks.
\newblock {\em CoRR} {\bf 2016}, {\em abs/1603.07292}.

\bibitem[Thrun \em{et~al.}(2021)Thrun, Ultsch, and Breuer]{make3010009}
Thrun, M.C.; Ultsch, A.; Breuer, L.
\newblock Explainable AI Framework for Multivariate Hydrochemical Time Series.
\newblock {\em Machine Learning and Knowledge Extraction} {\bf 2021}, {\em
  3},~170--204.
\newblock
  doi:{\changeurlcolor{black}\href{https://doi.org/10.3390/make3010009}{\detokenize{10.3390/make3010009}}}.

\bibitem[Bach \em{et~al.}(2015)Bach, Binder, Montavon, Klauschen, M{\"u}ller,
  and Samek]{bach2015pixel}
Bach, S.; Binder, A.; Montavon, G.; Klauschen, F.; M{\"u}ller, K.R.; Samek, W.
\newblock On pixel-wise explanations for non-linear classifier decisions by
  layer-wise relevance propagation.
\newblock {\em PloS one} {\bf 2015}, {\em 10},~e0130140.

\bibitem[Baehrens \em{et~al.}(2010)Baehrens, Schroeter, Harmeling, Kawanabe,
  Hansen, and M{\"u}ller]{baehrens2010explain}
Baehrens, D.; Schroeter, T.; Harmeling, S.; Kawanabe, M.; Hansen, K.;
  M{\"u}ller, K.R.
\newblock How to explain individual classification decisions.
\newblock {\em The Journal of Machine Learning Research} {\bf 2010}, {\em
  11},~1803--1831.

\bibitem[Arras \em{et~al.}(2017)Arras, Horn, Montavon, M{\"u}ller, and
  Samek]{arras2017relevant}
Arras, L.; Horn, F.; Montavon, G.; M{\"u}ller, K.R.; Samek, W.
\newblock " What is relevant in a text document?": An interpretable machine
  learning approach.
\newblock {\em PloS one} {\bf 2017}, {\em 12},~e0181142.

\bibitem[Hois \em{et~al.}(2019)Hois, Theofanou-Fuelbier, and Junk]{Hois_2019}
Hois, J.; Theofanou-Fuelbier, D.; Junk, A.J.
\newblock How to Achieve Explainability and Transparency in Human {AI}
  Interaction.
\newblock  HCI International 2019 - Posters; Stephanidis, C., Ed.; Springer
  International Publishing: Cham,  2019; pp. 177--183.

\bibitem[Alber \em{et~al.}(2019)Alber, Lapuschkin, Seegerer, H{\"a}gele,
  Sch{\"u}tt, Montavon, Samek, M{\"u}ller, D{\"a}hne, and
  Kindermans]{alber2019innvestigate}
Alber, M.; Lapuschkin, S.; Seegerer, P.; H{\"a}gele, M.; Sch{\"u}tt, K.T.;
  Montavon, G.; Samek, W.; M{\"u}ller, K.R.; D{\"a}hne, S.; Kindermans, P.J.
\newblock i{NN}vestigate neural networks!
\newblock {\em Journal of Machine Learning Research} {\bf 2019}, {\em
  20},~1--8.

\bibitem[Nori \em{et~al.}(2019)Nori, Jenkins, Koch, and
  Caruana]{nori2019interpretml}
Nori, H.; Jenkins, S.; Koch, P.; Caruana, R.
\newblock InterpretML: A Unified Framework for Machine Learning
  Interpretability,  2019,
  \href{http://xxx.lanl.gov/abs/1909.09223}{{\normalfont
  [arXiv:cs.LG/1909.09223]}}.

\bibitem[Wu \em{et~al.}(2019)Wu, Brooks, Chen, Chen, Choudhury, Dukhan,
  Hazelwood, Isaac, Jia, Jia, et~al.]{wu2019fb}
Wu, C.J.; Brooks, D.; Chen, K.; Chen, D.; Choudhury, S.; Dukhan, M.; Hazelwood,
  K.; Isaac, E.; Jia, Y.; Jia, B.; others.
\newblock Machine learning at {Facebook}: Understanding inference at the edge.
\newblock  2019 IEEE International Symposium on High Performance Computer
  Architecture (HPCA). IEEE,  2019, pp. 331--344.

\bibitem[Sehgal and Kehtarnavaz(2019)]{make1010027}
Sehgal, A.; Kehtarnavaz, N.
\newblock Guidelines and Benchmarks for Deployment of Deep Learning Models on
  Smartphones as Real-Time Apps.
\newblock {\em Machine Learning and Knowledge Extraction} {\bf 2019}, {\em
  1},~450--465.
\newblock
  doi:{\changeurlcolor{black}\href{https://doi.org/10.3390/make1010027}{\detokenize{10.3390/make1010027}}}.

\bibitem[{Christidis} \em{et~al.}(2019){Christidis}, {Davies}, and
  {Moschoyiannis}]{workload}
{Christidis}, A.; {Davies}, R.; {Moschoyiannis}, S.
\newblock Serving Machine Learning Workloads in Resource Constrained
  Environments: a Serverless Deployment Example.
\newblock  2019 IEEE 12th Conference on Service-Oriented Computing and
  Applications (SOCA),  2019, pp. 55--63.
\newblock
  doi:{\changeurlcolor{black}\href{https://doi.org/10.1109/SOCA.2019.00016}{\detokenize{10.1109/SOCA.2019.00016}}}.

\bibitem[Wang and Deng(2018)]{Wang_2018}
Wang, M.; Deng, W.
\newblock Deep visual domain adaptation: A survey.
\newblock {\em Neurocomputing} {\bf 2018}, {\em 312},~135 -- 153.

\bibitem[Sugiyama \em{et~al.}(2007)Sugiyama, Krauledat, and
  M{\"u}ller]{sugiyama2007covariate}
Sugiyama, M.; Krauledat, M.; M{\"u}ller, K.R.
\newblock Covariate shift adaptation by importance weighted cross validation.
\newblock {\em Journal of Machine Learning Research} {\bf 2007}, {\em
  8},~985--1005.

\bibitem[Heckemann \em{et~al.}(2011)Heckemann, Gesell, Pfister, Berns,
  Schneider, and Trapp]{heckemann2011safe}
Heckemann, K.; Gesell, M.; Pfister, T.; Berns, K.; Schneider, K.; Trapp, M.
\newblock Safe automotive software.
\newblock  International Conference on Knowledge-Based and Intelligent
  Information and Engineering Systems. Springer,  2011, pp. 167--176.

\bibitem[Berkenkamp \em{et~al.}(2016)Berkenkamp, Moriconi, Schoellig, and
  Krause]{berkenkamp2016safe}
Berkenkamp, F.; Moriconi, R.; Schoellig, A.P.; Krause, A.
\newblock Safe learning of regions of attraction for uncertain, nonlinear
  systems with gaussian processes.
\newblock  2016 IEEE 55th Conference on Decision and Control (CDC). IEEE,
  2016, pp. 4661--4666.

\bibitem[Derakhshan \em{et~al.}(2019)Derakhshan, Mahdiraji, Rabl, and
  Markl]{derakhshan2019continuous}
Derakhshan, B.; Mahdiraji, A.R.; Rabl, T.; Markl, V.
\newblock Continuous Deployment of Machine Learning Pipelines.
\newblock  EDBT,  2019, pp. 397--408.

\bibitem[Fehling \em{et~al.}(2014)Fehling, Leymann, Retter, Schupeck, and
  Arbitter]{CloudComputingPatterns2014}
Fehling, C.; Leymann, F.; Retter, R.; Schupeck, W.; Arbitter, P.
\newblock {\em Cloud Computing Patterns: Fundamentals to Design, Build, and
  Manage Cloud Applications}; Springer,  2014.
\newblock
  doi:{\changeurlcolor{black}\href{https://doi.org/10.1007/978-3-7091-1568-8}{\detokenize{10.1007/978-3-7091-1568-8}}}.

\bibitem[Ghanta \em{et~al.}(2018)Ghanta, Subramanian, Sundararaman, Khermosh,
  Sridhar, Arteaga, Luo, Das, and Talagala]{ghanta2018interpretability}
Ghanta, S.; Subramanian, S.; Sundararaman, S.; Khermosh, L.; Sridhar, V.;
  Arteaga, D.; Luo, Q.; Das, D.; Talagala, N.
\newblock Interpretability and Reproducability in Production Machine Learning
  Applications.
\newblock  2018 17th IEEE International Conference on Machine Learning and
  Applications (ICMLA). IEEE,  2018, pp. 658--664.

\end{thebibliography}

\end{document}